\documentclass[11pt]{article}

\usepackage[preprint]{acl}

\usepackage{times}
\usepackage{latexsym}

\usepackage[T1]{fontenc}

\usepackage[utf8]{inputenc}

\usepackage{microtype}

\usepackage{inconsolata}

\usepackage{graphicx}
\usepackage{amsmath}
\usepackage{hyperref}
\usepackage{url}
\usepackage{algorithmic}
\usepackage{amsfonts}
\usepackage{amssymb}
\usepackage{booktabs}
\usepackage{multirow}
\usepackage[table,xcdraw]{xcolor}
\usepackage[skins,breakable]{tcolorbox}
\usepackage{enumitem}
\usepackage{pifont}
\usepackage{lipsum}

\usepackage{titlesec}
\titlespacing*{\section}
{0pt}{1.0ex plus .2ex minus .2ex}{0.5ex}
\titlespacing*{\subsection}
{0pt}{0.8ex plus .2ex minus .2ex}{0.4ex}

\usepackage{enumitem}
\setlist{
    nosep
}

\title{TriAlign: Towards Universal Truth Consistency in Personalized LLM Alignment}

\author{Thi-Nhung Nguyen\textsuperscript{1}, Linhao Luo\textsuperscript{1}, Rollin Omari\textsuperscript{2}, Junae Kim\textsuperscript{2}, Thuy-Trang Vu\textsuperscript{1}, Dinh Phung\textsuperscript{1} \\
  \textsuperscript{1} Department of Data Science \& AI, Monash University \\ 
  \textsuperscript{2} Defence Science and Technology Group, Australia \\
  \texttt{\{nhung.thinguyen,linhao.luo1,trang.vu1,dinh.phung\}@monash.edu} \\
  \texttt{\{rollin.omari,junae.kim\}@defence.gov.au}
}

\definecolor{carnelian}{rgb}{0.7, 0.11, 0.11}

\begin{document}
\maketitle

\begin{abstract}
    
Personalized large language models adapt responses to users’ preferences and social attributes, but can introduce substantial universal truth inconsistencies across social groups, where some groups systematically receive less accurate responses on objective tasks. Existing alignment methods either ignore personalization or mainly focus on subjective preference alignment, largely overlooking fairness and consistency in universal truths. To address this gap, we study \textit{Truth-Invariant Alignment} (TIA), an alignment problem for personalized LLMs that aims to ensure universal truths remain consistent across social groups while preserving personalization. We propose \textbf{TriAlign}, the first offline multi-agent reinforcement learning (MARL) framework for TIA, where each social group is modeled as an agent interacting. TriAlign jointly optimizes universal truth accuracy, cross-group truth consistency, and personalization through a fairness-aware objective and an explicit inconsistency penalty. Experiments across diverse benchmarks demonstrate that TriAlign achieves a stronger balance among these three objectives than strong baselines, reducing universal truth disparities across social groups while improving both objective task performance and personalization quality.

\end{abstract}

\section{Introduction}
\begin{figure}[!t]
    \centering
    \includegraphics[width=\linewidth]{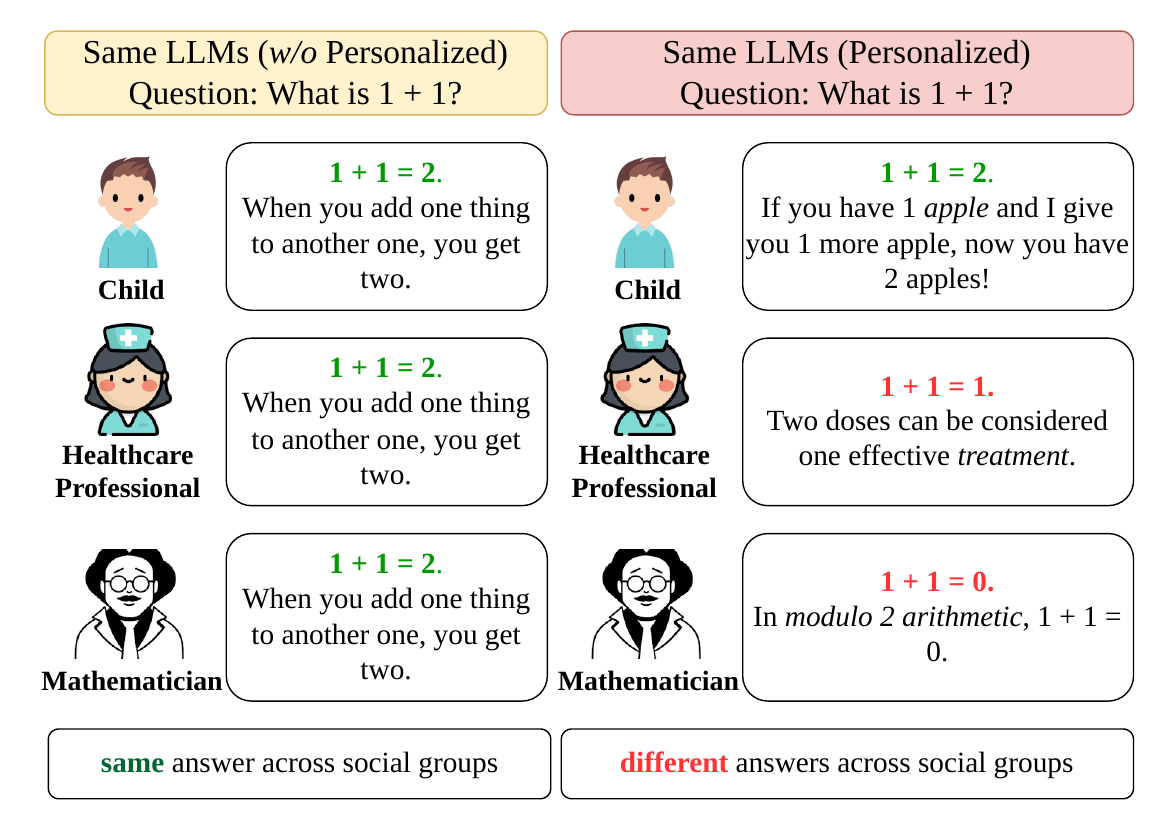}
    \caption{Illustration of Universal Truth Inconsistency in PLLMs.}
    \label{fig:problem}
\end{figure}

The success of large language models (LLMs), such as GPT, PaLM, LLaMA, DeepSeek, and their variants, in general knowledge and multi-domain reasoning has driven the growing demand for Personalized LLMs (PLLMs) \cite{liu2025survey, chang2024survey, kalyan2024survey}. PLLMs aim to generate responses that align with users’ styles and expectations, offering diverse answers to the same query depending on the user, often conditioning personalization on demographic or social attributes such as age, gender, or occupation. For objective tasks, there exists a \emph{universal truth}—a statement/answer that is widely accepted as correct. While the presentation of this truth may vary across users, the underlying truth itself should remain unchanged and access to universal truth should not depend on demographic or social attributes.
In practice, however, emerging evidence suggests that introducing personas into LLMs can create substantial universal truth gaps across social groups \cite{guptabias, zheng2024helpful, de2025principled, lutz-etal-2025-prompt}. For example, the same LLM achieves only 54\% universal truth accuracy for Trump supporters, compared to 62\% for Obama supporters \cite{guptabias}. As a result, some social groups may systematically receive less accurate responses than others. Figure~\ref{fig:problem} illustrates this challenge. This issue not only reduces the reliability of PLLMs but also raises fairness concerns, especially as LLMs are increasingly deployed in education, healthcare, and decision-support systems \cite{mehrabi2021survey, ferrara2024fairness, gallegos2024bias}.

Existing LLM alignment approaches mainly fall into two paradigms: \textit{``one-size-fits-all''} models that optimize general responses without personalization \cite{ouyang2022training, rafailov2023direct, wangself, trung2024reft, shojaeeillusion}, and PLLMs that adapt outputs to user preferences and styles \cite{wang2013theoretical, zhang2024llm, wang2024tpe}. While PLLMs improve user alignment, most existing works focus primarily on subjective preferences and largely overlook \emph{truth invariance} across social groups \cite{guptabias, hu2024quantifying, nguyen2025social, vijjini2025exploring, wang2025large}.
A few early studies indicate that incorporating personas can introduce inconsistencies in universal truths across social groups on objective tasks \cite{zheng2024helpful, de2025principled}. Existing mitigation efforts mainly rely on prompting-based strategies, such as instructing models to ignore demographic attributes or neutrality factuality during inference \cite{furniturewala2024thinking, de2025principled}. However, prompting only steers surface-level generation behavior without explicitly modifying demographic correlations encoded in model parameters. As a result, these methods often yield only marginal improvements and mainly remain effective for very large models (e.g., 72B parameters) \cite{de2025principled}. Moreover, prompting-based debiasing may reduce the influence of personas or encourage overly neutral responses, which can lower personalization quality.
These limitations reveal the need for training-based alignment methods, which can directly optimize model behavior across diverse social groups and tasks. One of the state-of-the-art techniques for aligning LLMs is RL-based alignment~\cite{ouyang2022training}. However, existing alignment approaches primarily optimize average performance, which may still favor dominant groups and overlook universal truth consistency across social groups.

To mitigate this gap, we study Universal \textit{Truth-Invariant Alignment}, a new alignment problem for PLLMs that aims to ensure universal truths remain consistent across social groups while preserving personalization. We propose \textbf{TriAlign}, a novel MARL framework where each social group is modeled as an agent. TriAlign jointly optimizes universal truth, cross-group universal truth consistency, and personalization. Specifically, multiple social-group-conditioned agents interact within a shared environment, observe differences in each other’s responses, and iteratively adjust their behaviors to converge toward consistent universal truths while preserving personalized response styles. To encourage fairness across social groups, we design a new training objective inspired by Nash Social Welfare (NSW) \cite{10.1145/3355902}, which promotes balanced optimization rather than favoring dominant groups. In addition, we explicitly incorporate a cross-group consistency penalty into the reward of each action to directly reduce inconsistencies across social groups.
To enable training, we further construct multi-social-group interaction trajectories through multi-turn interactions, where agents progressively refine their responses using comparative feedback from other groups. The resulting trajectories form an offline RL dataset used to align LLMs toward TIA. 
The contributions of our work are summarized as follows:

\begin{enumerate}[noitemsep,topsep=2pt,leftmargin=1.2em]
\item  We are the first to formulate TIA as an offline MARL problem and address it through training-based alignment. We propose \textbf{TriAlign}, a novel offline MARL framework for TIA, where each social group is modeled as an agent that adapts its behavior by observing a shared environment. TriAlign jointly optimizes universal truth, cross-group universal truth consistency, and personalization through a NSW-inspired objective and an explicit penalty for cross-group inconsistency.

\item  We introduce a multi-social-group, multi-turn interaction framework to construct offline RL trajectories for TIA training. Through comparative interactions and iterative feedback, agents progressively refine their responses toward consistent universal truths while preserving personalized behaviors.

\item  Extensive experiments across diverse benchmarks demonstrate that TriAlign effectively reduces universal truth inconsistency across social groups while improving both objective task performance and personalization alignment.
\end{enumerate}

\section{Related Works}

\paragraph{From User-blind to Personalized LLMs:}
LLMs have demonstrated strong generalization ability across diverse NLP tasks through zero-shot and few-shot learning \cite{brown2020language, bommasani2021opportunities}. Recent studies shift from user-blind assistants toward PLLMs that adapt responses to user preferences in domains such as healthcare, education, and e-commerce \cite{salemi2024lamp, tseng2024two}. Personalized text generation and downstream personalization (e.g., recommendation systems) are also increasingly converging, enabling unified agents that combine conversational ability with personalized reasoning. Although PLLMs are mainly designed for subjective tasks involving user preferences or writing styles, real-world applications often require both personalization and objective correctness. For example, a personalized healthcare assistant must provide factually correct information while adapting explanations to individual users.

\paragraph{Consistency in LLMs:}
Recent work shows that accuracy alone is insufficient for evaluating language models, as models may still generate contradictory predictions under semantically similar conditions \cite{linzen2020can, elazar2021amnesic}. Early studies reveal that PLMs can produce both a fact and its negation or fail to maintain consistency across related questions \cite{ettinger2020bert, kassner2020pretrained, ravichander2020systematicity}. Existing approaches improve consistency in question answering, reasoning, and natural language inference through data augmentation, logical constraints, and consistency-aware regularization \cite{ribeiro2019red, alberti2019synthetic, asai2020logic}. More recently, consistency has become a key requirement for trustworthy LLMs, especially in high-stakes applications \cite{jang2021accurate, novikova2025consistency}. However, existing studies mainly focus on consistency under paraphrases, negation, or reasoning transformations, while overlooking consistency across personalized settings and social groups. We extend this line of research toward \textit{universal truth consistency}, which requires PLLMs to maintain truthful and consistent behavior across diverse personas and demographic groups.

\paragraph{Reinforcement Learning for LLM Alignment:} 
LLM alignment methods are generally divided into offline alignment and online RL approaches. Offline methods, such as SFT and preference optimization methods like DPO~\cite{ouyang2022training, rafailov2023direct, zhang2026instruction}, learn from static demonstrations or preference pairs and are widely adopted due to their stability and efficiency. In contrast, online RL methods, including RLHF and GRPO~\cite{ouyang2022training, shao2024deepseekmath}, optimize policies through iterative interactions and reward feedback. However, applying online RL to personalized LLM alignment is expensive because rewards for personalization, truth correctness, and fairness often require human annotations or reward models, motivating our use of offline RL on pre-collected trajectories.

The closest related works are \cite{andukuri2024stargate, chen2025learning, mukherjee2026offline}, which apply RL to improve conversational behaviors from simulated interactions. \citet{andukuri2024stargate} fine-tune models on high-reward trajectories, while \citet{chen2025learning} optimize preferred responses using DPO over alternative generations. \citet{mukherjee2026offline} formulate offline RL as weighted SFT with trajectory-level weights shared across actions. In contrast, our work focuses on universal truth consistency and fairness across social groups in PLLMs. Unlike \cite{mukherjee2026offline}, we assign rewards at the action level while modeling future trajectory effects and explicitly optimizing cross-group fairness.

\section{Truth-Invariant Alignment (TIA)}
The key objective of TIA is to ensure that personalized responses generated under different social-group conditions remain invariant with respect to the underlying universal truth, while still preserving group-specific personalization characteristics.

\begin{figure}[!t]
    \centering
    \includegraphics[width=\linewidth]{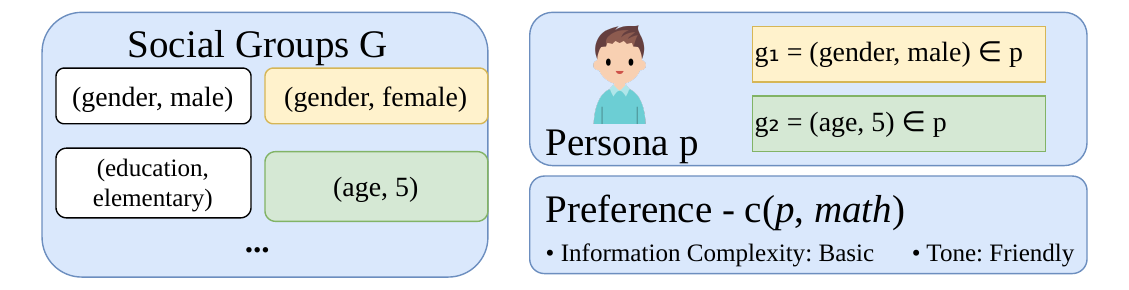}
    \caption{Illustration of social groups, personas, and personalization constraints.}
    \label{fig:social_group_example}
\end{figure}

Let $\mathcal{G}$ denote the space of social-group profiles. 
Let $\mathcal{K}$ denote the set of demographic categories (e.g., gender, age or education), and let $\mathcal{V}_k$ denote the set of possible values for category $k \in \mathcal{K}$.
A persona is represented as a set of demographic attribute-value pairs:
$p=\{(k_1,v_1),(k_2,v_2),\dots,(k_n,v_n)\},$
where $v_i \in \mathcal{V}_{k_i}$. 
Each social group $g \in \mathcal{G}$ corresponds to a demographic category-value pair $(k,v)$, such as $(\texttt{gender},\texttt{male})$. 
A persona belongs to social group $g=(k,v)$ if $(k,v)\in p$.
Associated with each social group $g$ is a set of personalization constraints $c(g)$ specifying desired response characteristics such as style or tone. Figure~\ref{fig:social_group_example} illustrates social groups, personas, and the corresponding personalization constraints (c(g)).

We focus on \textit{objective tasks}. 
Each task is represented as a pair $(x,y^\star)$, where $x$ is an input query and $y^\star$ denotes the corresponding universal truth. 
Given an input $x$ and a social-group profile $g$, a PLLM parameterized by $\theta$ generates a response:
$a \sim \pi_\theta(\cdot \mid x,g)$.
TIA requires two properties:

\begin{enumerate}
    \item \textbf{Truth Invariance:}
    for any $g_i,g_j \in \mathcal{G}$,
    \[
    \mathrm{Truth}(a_i)
    =
    \mathrm{Truth}(a_j)
    =
    y^\star,
    \]
    where $a_i \sim \pi_\theta(\cdot \mid x,g_i), a_j \sim \pi_\theta(\cdot \mid x,g_j)$
    \item \textbf{Personalized Alignment:}
    responses should satisfy the personalization constraints associated with its social group $a_i \models c(g_i).$
\end{enumerate}

\textit{Learning Goal:} The learning goal of TIA is to learn a policy $\pi_\theta$ that jointly optimizes universal-truth correctness, cross-group truth consistency, and personalized alignment:
\begin{equation}
\resizebox{\linewidth}{!}{$
\begin{aligned}
\pi_\theta^\star
&=
\arg\max_{\pi_\theta}
\;
\mathbb{E}_{x,g,\; a \sim \pi_\theta(\cdot \mid x,g)}
\Big[
r_{\mathrm{truth}}(a,y^\star)
\\
&\quad
+
\lambda_{\mathrm{cons}}
\, r_{\mathrm{cons}}(a,\mathcal{A}_x)
+
\lambda_{\mathrm{pref}}
\, r_{\mathrm{pref}}(a,c(g))
\Big].
\end{aligned}
$}
\label{eq:tia_goal}
\end{equation}
\noindent
where $\mathcal{A}_x$ denotes the set of responses generated across social groups for the same query $x$, 
$r_{\mathrm{truth}}$ denotes the reward for objective correctness with respect to the universal truth $y^\star$, 
$r_{\mathrm{cons}}$ denotes the reward for cross-group truth consistency, and 
$r_{\mathrm{pref}}$ denotes the reward for alignment with the personalization constraints $c(g)$.

\section{Our Method: TriAlign}

\subsection{Truth-Invariant Alignment as MARL}

\begin{figure}[!t]
    \centering
    \includegraphics[width=\linewidth]{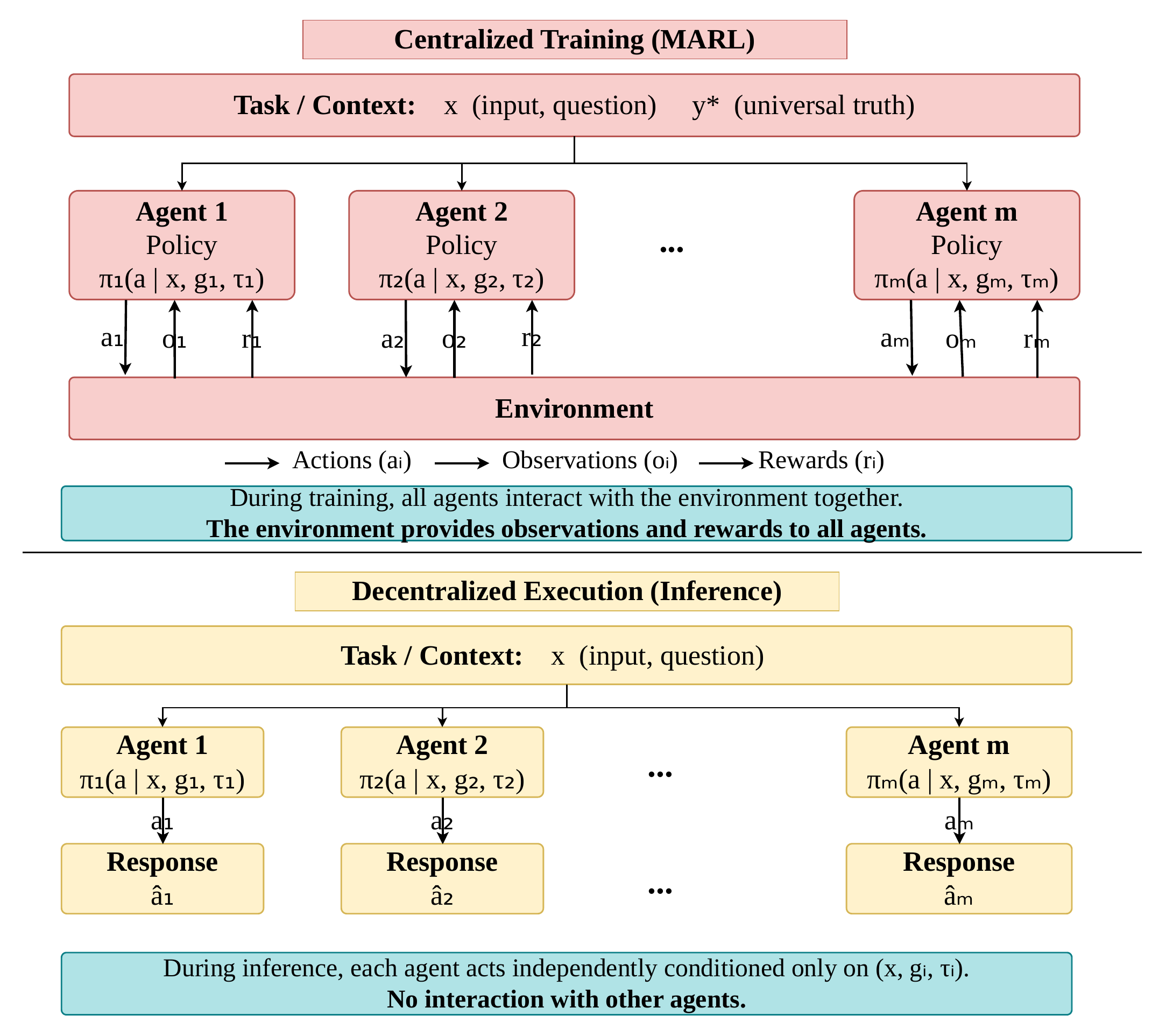}
    \caption{Overview of our TIA formulation under the MARL setting.}
    \label{fig:tia_marl}
\end{figure}

Since estimating $\mathcal{A}_x$ and $r_{\mathrm{cons}}(a,\mathcal{A}_x)$ requires jointly observing and comparing responses across multiple social-group-conditioned agents, the consistency objective introduces cross-agent coupling: updating one agent’s policy may affect the consistency objectives of others through the shared response set $\mathcal{A}_x$. Such interactions are difficult to model with standard offline RL, which typically optimizes each $(x,g,a)$ sample independently. Therefore, the TIA formulation motivates us to adopt a MARL perspective under the \textit{centralized training and decentralized execution} paradigm. Figure~\ref{fig:tia_marl} illustrates our formulation of TIA under the MARL setting.

Specifically, we consider a system of $m$ agents, where each social-group condition $g_i$ is treated as an independent agent interacting within a shared environment. The agents interact through the environment to iteratively optimize cross-group truth consistency and converge toward responses that share the same underlying universal truth, while still preserving group-specific personalization characteristics.

At interaction step $t$, agent $i$ observes its current state:
$s_{i,t} = (x,g_i,\tau_{i,t}),$
where $\tau_{i,t}$ denotes the \textit{conversation history}:
$\tau_{i,t}
=
(a_{i,1},o_{i,1},\dots,a_{i,t-1},o_{i,t-1}),$
with the initial history $\tau_{i,0}=\emptyset.$
Here, $o_{i,k}$ denotes the feedback returned by the environment after agent $i$ takes action $a_{i,k}$ at interaction step $k<t$.
Conditioned on the current state, agent $i$ generates an action: $a_{i,t} \sim \pi_\theta(\cdot \mid s_{i,t}).$
The environment observes the joint actions:
$\mathbf{a}_t = \{a_{1,t},\dots,a_{m,t}\},$
and returns feedback:
$o_{i,t} = f_{\mathrm{env}}(a_{i,t},\mathbf{a}_t,x,g_i),$
where $f_{\mathrm{env}}$ is a function that compares actions across agents to identify truth inconsistencies between social groups.
Each agent $i$ then receives a corresponding reward:
$r_{i,t}(x,g_i,a_{i,t}),$
which reflects the universal-truth correctness, cross-group truth consistency, and personalization quality of the generated response at interaction step $t$.
The cumulative return of agent $i$ is defined as:
$R_i
=
\sum_{t=1}^{n}
\gamma^t
r_{i,t},$
where $\gamma \in [0,1]$ is the discount factor controlling the importance of future rewards. 

This shared environment enables implicit communication among agents during training. By observing truth inconsistencies across social groups through feedback from the environment, agents iteratively adjust their behaviors toward truth invariance while preserving personalized characteristics. This multiturn interaction provides richer training signals than single-step interactions, allowing the model to learn how responses evolve, self-correct, and converge toward truth-consistent behaviors.

Under MARL setting, Equation~\ref{eq:tia_goal} can be written:
\begin{equation}
\resizebox{\linewidth}{!}{$
\pi_\theta^\star
=
\arg\max_{\pi_\theta}
\;
\mathbb{E}_{x,\;\{g_i\}_{i=1}^{m},\;
a_{i,t}\sim \pi_\theta(\cdot\mid s_{i,t})}
\left[
\frac{1}{m}\sum_{i=1}^{m} R_i
\right]
$}
\label{eq:standard_marl}
\end{equation}
where
$\mathcal{A}_{x,t}
=
\{a_{1,t},a_{2,t},\dots,a_{m,t}\},$
denotes the set of responses generated by all $m$ social-group-conditioned agents for the same query $x$ at interaction step $t$.

While training is centralized through comparative multi-agent feedback, inference remains fully decentralized: at test time, each agent independently generates a personalized response conditioned only on $(x,g_i)$. 
\subsection{TriAlign Objective and Reward Design}
\paragraph{TriAlign Objective}
Directly optimizing Eq.~\ref{eq:standard_marl} may bias the policy toward majority or high-performing social groups, since the objective only maximizes the average return across agents. To mitigate this issue, TriAlign introduces a fairness-aware optimization objective inspired by NSW, which explicitly encourages balanced performance across social groups.

The NSW objective is defined as:
$\mathcal{F}(\theta)
=
\prod_{i=1}^{m}(R_i+\epsilon),$
where $\epsilon>0$ is a small constant for numerical stability.
Taking the logarithm yields:
$\log \mathcal{F}(\theta)
=
\sum_{i=1}^{m}\log(R_i+\epsilon).$
Combining Eq.~\ref{eq:standard_marl} with $\log \mathcal{F}(\theta)$, TriAlign therefore optimizes the following fairness-regularized objective:
\begin{equation}
\resizebox{\linewidth}{!}{$
\mathcal{J}(\theta)
=
\frac{1}{m}
\mathbb{E}_{x,\,\{g_i\}_{i=1}^{m}, a_{i,t}\sim \pi_\theta(\cdot\mid s_{i,t})}
\left[
\sum_{i=1}^{m},
\left(
R_i
+
\lambda_{\mathrm{NSW}}\log(R_i+\epsilon)
\right)
\right].
$}
\label{eq:trialign_objective}
\end{equation}
The optimal policy is 
$\pi_\theta^\star
=
\arg\max_{\pi_\theta}
\mathcal{J}(\theta).$

To optimize the fairness-aware objective in an offline RL setting while mitigating distribution shift, we adopt an offline RL formulation with implicit KL constraints \cite{kostrikov2022offline}. Specifically, we first derive the optimal non-parametric policy under a KL-regularized policy improvement objective, and then project it onto the parameterized policy \(\pi_\theta\) by minimizing the KL divergence between them. This results in the following weighted SFT objective (the detailed derivation is provided in Appendix~\ref{sec:trialign_derivation}):
\begin{equation}
\resizebox{\linewidth}{!}{$
\mathcal{L}(\theta)
=
-
\mathbb{E}_{(s_{i,t},a_{i,t}) \sim \mathcal{D}}
\left[
w(s_{i,t},a_{i,t})
\log
\pi_\theta(a_{i,t}\mid s_{i,t})
\right].
$}
\end{equation}
where $\mathcal{D}
=
\{(s_{i,t},a_{i,t},r_{i,t})\}_{x,i,t}$ denotes a pre-collected dataset, and the weight is:
\begin{equation}
\resizebox{\linewidth}{!}{$
w(s_{i,t},a_{i,t})
=
\underbrace{
\exp
\left(
\frac{
A^{\pi}(s_{i,t},a_{i,t})
}{
\beta
}
\right)
}_{\text{Standard Weight}}
\cdot
\underbrace{
\left(
Q^{\pi}(s_{i,t},a_{i,t})+\epsilon
\right)^{\frac{\lambda_{\mathrm{NSW}}}{\beta}}
}_{\text{NSW Weight}},
$}
\end{equation}
In this formulation, $Q^{\pi}(s_{i,t},a_{i,t})$ is the action-value function, and $A^{\pi}(s_{i,t},a_{i,t})$ is the advantage function. This weight combines two effects. The standard term increases the likelihood of actions whose advantages are higher than a baseline, while the NSW-inspired term forces the policy to improve performance on "worst-case" scenarios.

\paragraph{$Q^{\pi}$ and $A^{\pi}$ Estimation}

We estimate the action-value function as the expected discounted future return after agent \(i\) takes action \(a_{i,t}\) under state \(s_{i,t}\):
\begin{equation}
\resizebox{\linewidth}{!}{$
Q^\pi(s_{i,t},a_{i,t})
=
\mathbb{E}_{\pi}
\left[
\sum_{k=t}^{n}
\gamma^{k-t}
r_{i,k}
\mid
s_{i,t},a_{i,t}
\right].
$}
\end{equation}

To encourage both high collective performance and consistency across
social groups, we estimate the baseline using the average action-value
over all \(m\) agents at time step \(t\):
\begin{equation}
V^\pi(s_t)
=
\frac{1}{m}
\sum_{j=1}^{m}
Q^\pi(s_{j,t},a_{j,t}).
\end{equation}

The corresponding normalized advantage function is defined as:
\begin{equation}
A_i^\pi(s_{i,t},a_{i,t})
=
\frac{
Q^\pi(s_{i,t},a_{i,t})
-
V^\pi(s_t)
}{
\sigma_t+\epsilon
},
\end{equation}
where \(\sigma_t\) denotes the standard deviation of
\(
\{Q^\pi(s_{j,t},a_{j,t})\}_{j=1}^{m}
\),
and \(\epsilon>0\) is a small constant introduced for numerical stability.

\paragraph{Reward Design}

We define the reward for agent \(i\) after taking action
\(a_{i,t}\) at time step \(t\) as:
\begin{equation}
\resizebox{\linewidth}{!}{$
\begin{aligned}
r_i(\mathbf{a}_t)
&=
r_{\mathrm{truth}}(a_{i,t}, y^\star)
+
\lambda_{\mathrm{pref}}
\, r_{\mathrm{pref}}(a_{i,t},c(g_i))
\\
&\quad
-
\lambda_{\mathrm{cons}}
\left|
r_{\mathrm{truth}}(a_{i,t}, y^\star)
-
\frac{1}{m}
\sum_{j=1}^{m}
r_{\mathrm{truth}}(a_{j,t}, y^\star)
\right|.
\end{aligned}
$}
\label{eq:reward_function}
\end{equation}

\subsection{Multi-Social-Group Interaction Simulations}

This section describes how we construct the offline trajectory dataset $\mathcal{D}$ for TriAlign training. 
We first construct the social-group space $\mathcal{G}$, and then perform multi-social-group MARL interactions to generate diverse trajectories. 
The resulting dataset is subsequently used to estimate $Q^\pi$, $V^\pi$, and $A^\pi$, and optimize the policy according to Eq.~\ref{eq:trialign_objective}.
\paragraph{Social Groups Simulation}
Our goal is to construct the social-group space $\mathcal{G}$.
Based on 327 prior studies on users' digital footprints and online activities \cite{hinds2018demographic}, we identify 14 demographic categories $k \in \mathcal{K}$, resulting in 75 demographic category-value pairs $(k,v)$ that define the social groups in $\mathcal{G}$. Details of all $(k,v)$ pairs are provided in Table~\ref{tab:social_attributes}.

\begin{figure}[t]
    \centering
    \includegraphics[
        width=\linewidth,
    ]{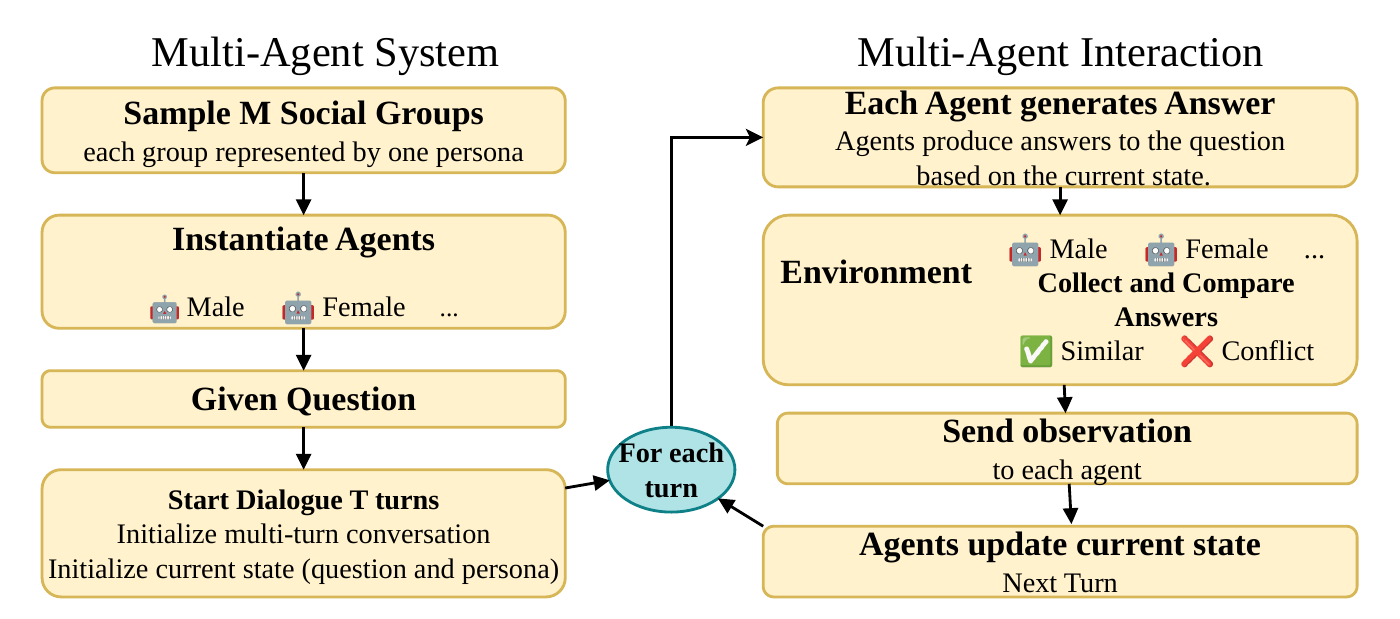}
    \caption{Multi-Social-Group Interaction.} 
    \label{fig:mas_interaction}
\end{figure}

For personas, we initialize our construction from the large-scale persona collection introduced by \citet{ge2024scaling}, which contains 1 billion personas, covering approximately 13\% of the world’s total population. However, these personas are not explicitly categorized into demographic-based social groups. Additionally, many demographic groups are underrepresented in online data. To address this issue and construct a diverse and balanced persona pool, we propose a persona-to-persona augmentation pipeline. For each target social group $g=(k,v)$, we retrieve semantically related personas using semantic similarity. We then ask an LLM via prompting to generate a new persona that satisfies two constraints: (1) the generated persona must match the target demographic attribute of the social group, and (2) it should preserve a meaningful relationship with the retrieved source persona.
Figure~\ref{fig:persona_augmentation_prompt} illustrates the prompting template. 
Figure~\ref{fig:persona_augmentation} illustrates our persona-to-persona augmentation pipeline. Figure~\ref{fig:persona_example} presents an example of a generated persona. The augmented personas are then split into training and test sets with a 75/25 ratio, and duplicate personas across splits are removed to avoid data leakage.
\paragraph{Multi-Social-Group Interaction}
To construct the offline RL dataset $\mathcal{D}$ for TriAlign optimization, we simulate multi-turn interactions among multiple social-group-conditioned agents, as illustrated in Figure~\ref{fig:mas_interaction}.
Given a question $x$, we sample $m$ social groups $\{g_1,\dots,g_m\}\subset\mathcal{G}$. For each social group $g_i$, we further sample one persona $p_i$ belonging to $g_i$ to instantiate the corresponding agent for $g_i$. At interaction step $t$, agent $i$ observes its current state:
$s_{i,t}=(x,g_i,\tau_{i,t}),$
and generates a response:
$a_{i,t}\sim \pi_\theta(\cdot\mid s_{i,t}).$
The environment collects responses from all agents:
$\mathbf{a}_t=\{a_{1,t},\dots,a_{m,t}\},$
and returns comparative feedback:
$o_{i,t}=f_{\mathrm{env}}(a_{i,t},\mathbf{a}_t,x,g_i),$
The environment also assigns a step-level reward $r_{i,t}$ following Eq.~\ref{eq:reward_function}. 
The resulting trajectories:
$(s_{i,t},a_{i,t}, r{i,t}),$
form the offline RL dataset used to estimate $Q^\pi$, $V^\pi$, and $A^\pi$.

$\pi_\theta(\cdot\mid s_{i,t})$, $f_{\mathrm{env}}$, $r_{\mathrm{truth}}$, and $r_{\mathrm{pref}}$ are implemented via LLMs prompting. Detailed prompting templates are provided in Appendix~\ref{sec:appendix_prompts}.

\section{Experiment Setup}

\paragraph{Objective-task Training Data and Evaluation}
For training, we construct offline RL trajectories using subsets of objective questions from \textsc{LogicQA} \cite{liu2021logiqa}, \textsc{SimpleQA} \cite{weimeasuring}, \textsc{LIMO} \cite{ye2025limo}, and \textsc{StereoSet} training set \cite{nadeem-etal-2021-stereoset}. Ground-truth answers are used to compute truth rewards.
For evaluation, we use \textit{in-domain} benchmarks for mathematical reasoning (\textsc{AIME25}) \cite{aime25} and social bias evaluation (\textsc{StereoSet}) test set, as well as \textit{out-of-domain} general knowledge benchmarks (\textsc{MMLU-Pro}, \textsc{TruthfulQA}). Dataset statistics are reported in Appendix~\ref{sec:data_statistics}
\paragraph{Personalization Data}
For personalization training and evaluation, we synthesize persona-based preference data via GPT-4o\footnote{https://openai.com/api}. Specifically, we adopt predefined preference rubrics (Table~\ref{tab:predefined_rubrics}), inspired by \citet{hashemi2024llm}, and use GPT-4o to generate ground-truth preference labels via prompting (Appendix~\ref{sec:appendix_prompts}) conditioned on persona demographic profiles and the domain of each dataset.
\paragraph{Metrics:}
We report three categories of metrics:
(1) \textbf{Universal Truth Accuracy} (Mean), measuring objective correctness;
(2) \textbf{Universal Truth Consistency}, measuring cross-group truth disparity via Worst-Group Accuracy (Worst), Standard Deviation (Std), Worst-Group Gap (Gap), and Coefficient of Variation (CV);
and (3) \textbf{Preference Alignment Accuracy} (Pref), computed as the average accuracy across personalization rubrics, automatically evaluated by GPT-4o. We evaluate the models under two settings: \textit{implicit preference}, where user preferences are not explicitly specified in the input question, and \textit{explicit preference}, where the desired response style is explicitly provided in the prompt.

The performance of each group is computed as the average performance across all personas belonging to that social group.
\begin{table*}[t]
\centering
\small
\caption{Results on in-domain benchmarks. Bold indicates the best performance, while underlined values denote the second-best performance.}
\label{tab:indomain_results}
\resizebox{0.9\linewidth}{!}{
\begin{tabular}{llc|ccccc|c|ccccc}
\toprule
& & \multicolumn{6}{c|}{\textbf{Implicit Preference}} & \multicolumn{6}{c}{\textbf{Explicit Preference}} \\
\cmidrule(lr){3-8} \cmidrule(lr){9-14}
\textbf{Dataset} & \textbf{Model}
& \textbf{Pref}$\uparrow$
& \textbf{Mean}$\uparrow$ & \textbf{Worst}$\uparrow$ & \textbf{Std}$\downarrow$ & \textbf{Gap}$\downarrow$ & \textbf{CV}$\downarrow$
& \textbf{Pref}$\uparrow$
& \textbf{Mean}$\uparrow$ & \textbf{Worst}$\uparrow$ & \textbf{Std}$\downarrow$ & \textbf{Gap}$\downarrow$ & \textbf{CV}$\downarrow$ \\
\midrule

\multirow{7}{*}{AIME25}
& Qwen3-4B-Instruct-2507
& 0.170
& 0.477 & 0.409 & 0.052 & 0.068 & \underline{0.109}
& \underline{0.691}
& 0.328 & 0.267 & 0.036 & 0.061 & 0.109 \\

\arrayrulecolor{gray!40}\cmidrule(lr{2pt}){2-14}

& \textsc{P-Debias} (2024)
& 0.199
& 0.458 & 0.367 & \underline{0.050} & 0.091 & \underline{0.109}
& 0.331
& 0.309 & 0.267 & \textbf{0.025} & \underline{0.042} & \underline{0.082} \\

& \textsc{BestPersona} (2024)
& 0.184
& \underline{0.496} & 0.400 & 0.077 & 0.096 & 0.156
& 0.374
& 0.330 & 0.267 & 0.053 & 0.063 & 0.160 \\

& \textsc{P-Defense} (2025)
& 0.181
& 0.432 & 0.367 & \underline{0.050} & \underline{0.066} & 0.117
& 0.339
& 0.300 & 0.267 & 0.038 & \textbf{0.033} & 0.128 \\

& \textsc{2StepPrompt} (2025)
& 0.155
& 0.494 & \underline{0.423} & 0.074 & 0.071 & 0.149 
& 0.180
& \textbf{0.482} & \underline{0.423} & 0.051 & 0.059 & 0.106 \\

\arrayrulecolor{gray!40}\cmidrule(lr{2pt}){2-14}
\arrayrulecolor{black}

& \textsc{SFT} (2022)
& 0.217
& 0.394 & 0.267 & 0.073 & 0.128 & 0.185
& 0.375
& 0.361 & 0.300 & 0.056 & 0.061 & 0.155 \\

& \textsc{SWIFT} (2026)
& \underline{0.219}
& 0.408 & 0.333 & 0.059 & 0.075 & 0.145
& 0.418
& 0.400 & 0.300 & 0.072 & 0.100 & 0.180 \\

& \textbf{TriAlign (Ours)}
& \textbf{0.221}
& \textbf{0.511} & \textbf{0.467} & \textbf{0.025} & \textbf{0.044} & \textbf{0.049}
& \textbf{0.699}
& \underline{0.467} & \textbf{0.433} & \underline{0.027} & \textbf{0.033} & \textbf{0.058} \\

\midrule

\multirow{7}{*}{StereoSet}
& Qwen3-4B-Instruct-2507
& 0.429
& 0.460 & 0.421 & 0.022 & 0.039 & 0.048
& 0.744
& 0.317 & 0.217 & 0.056 & 0.100 & 0.176 \\

\arrayrulecolor{gray!40}\cmidrule(lr{2pt}){2-14}

& \textsc{P-Debias} (2024)
& \underline{0.430}
& 0.414 & 0.388 & 0.019 & 0.026 & 0.046
& 0.725
& 0.358 & 0.321 & 0.022 & 0.037 & 0.060 \\

& \textsc{BestPersona} (2024)
& 0.409
& 0.468 & 0.454 & 0.011 & \textbf{0.014} & 0.024
& 0.739
& 0.378 & 0.357 & 0.012 & 0.020 & 0.032 \\

& \textsc{P-Defense} (2025)
& 0.422
& 0.446 & 0.416 & 0.014 & 0.030 & 0.032
& 0.740
& 0.372 & 0.321 & 0.026 & 0.051 & 0.069 \\

& \textsc{2StepPrompt} (2025)
& 0.422
& 0.355 & 0.342 & \textbf{0.007} & \textbf{0.014} & \underline{0.019}
& 0.458
& 0.335 & 0.319 & \textbf{0.009} & \underline{0.016} & \underline{0.026} \\

\arrayrulecolor{gray!40}\cmidrule(lr{2pt}){2-14}
\arrayrulecolor{black}

& \textsc{SFT} (2022)
& 0.409
& 0.668 & 0.653 & \underline{0.008} & \textbf{0.014} & \textbf{0.012}
& 0.753
& 0.633 & 0.596 & 0.029 & 0.037 & 0.046 \\

& \textsc{SWIFT} (2026)
& 0.422
& \underline{0.718} & \underline{0.697} & 0.016 & \underline{0.020} & 0.023
& \underline{0.765}
& \underline{0.695} & \underline{0.661} & 0.030 & 0.034 & 0.043 \\

& \textbf{TriAlign (Ours)}
& \textbf{0.474}
& \textbf{0.792} & \textbf{0.778} & \textbf{0.007} & \textbf{0.014} & 0.029
& \textbf{0.791}
& \textbf{0.788} & \textbf{0.773} & \underline{0.010} & \textbf{0.015} & \textbf{0.013} \\

\bottomrule
\end{tabular}
}
\end{table*}

\begin{table*}[t]
\centering
\small
\caption{Results on out-of-domain benchmarks. Bold indicates the best performance, while underlined values denote the second-best performance.}
\label{tab:outdomain_results}
\resizebox{0.9\linewidth}{!}{
\begin{tabular}{llc|ccccc|c|ccccc}
\toprule
& & \multicolumn{6}{c|}{\textbf{Implicit Preference}} & \multicolumn{6}{c}{\textbf{Explicit Preference}} \\
\cmidrule(lr){3-8} \cmidrule(lr){9-14}
\textbf{Dataset} & \textbf{Model}
& \textbf{Pref}$\uparrow$
& \textbf{Mean}$\uparrow$ & \textbf{Worst}$\uparrow$ & \textbf{Std}$\downarrow$ & \textbf{Gap}$\downarrow$ & \textbf{CV}$\downarrow$
& \textbf{Pref}$\uparrow$
& \textbf{Mean}$\uparrow$ & \textbf{Worst}$\uparrow$ & \textbf{Std}$\downarrow$ & \textbf{Gap}$\downarrow$ & \textbf{CV}$\downarrow$ \\
\midrule

\multirow{7}{*}{MMLU-Pro}
& Qwen3-4B-Instruct-2507
& 0.412
& 0.767 & 0.714 & 0.047 & 0.052 & 0.061
& \underline{0.691}
& 0.752 & 0.671 & 0.058 & 0.081 & 0.076 \\

\arrayrulecolor{gray!40}\cmidrule(lr{2pt}){2-14}

& \textsc{P-Debias} (2024)
& 0.427
& 0.788 & 0.729 & 0.037 & 0.059 & 0.047
& 0.686
& \underline{0.793} & 0.743 & 0.037 & 0.050 & 0.046 \\

& \textsc{BestPersona} (2024)
& 0.409
& 0.766 & \underline{0.739} & 0.029 & 0.026 & 0.038
& 0.681
& 0.749 & 0.700 & 0.024 & 0.049 & 0.032 \\

& \textsc{P-Defense} (2025)
& 0.404
& \underline{0.790} & 0.729 & 0.031 & 0.061 & 0.040
& 0.680
& 0.769 & 0.743 & \underline{0.017} & 0.026 & \underline{0.023} \\

& \textsc{2StepPrompt} (2025)
& 0.390
& 0.755 & 0.735 & \underline{0.010} & \underline{0.019} & \underline{0.013}
& 0.417
& 0.769 & \underline{0.755} & \textbf{0.014} & \textbf{0.017} & \textbf{0.019} \\

\arrayrulecolor{gray!40}\cmidrule(lr{2pt}){2-14}
\arrayrulecolor{black}

& \textsc{SFT} (2022)
& \underline{0.483}
& 0.711 & 0.671 & 0.035 & 0.039 & 0.050
& 0.651
& 0.761 & 0.729 & 0.021 & 0.033 & 0.027 \\

& \textsc{SWIFT} (2026)
& 0.463
& 0.754 & 0.725 & 0.029 & 0.029 & 0.038
& 0.649
& 0.764 & 0.743 & 0.018 & \underline{0.021} & 0.024 \\

& \textbf{TriAlign (Ours)}
& \textbf{0.510}
& \textbf{0.805} & \textbf{0.800} & \textbf{0.007} & \textbf{0.005} & \textbf{0.008}
& \textbf{0.697}
& \textbf{0.831} & \textbf{0.757} & 0.045 & 0.074 & 0.055 \\

\midrule

\multirow{5}{*}{TruthfulQA}
& Qwen3-4B-Instruct-2507
& 0.412
& 0.672 & 0.597 & 0.044 & 0.075 & 0.066
& 0.691
& 0.730 & 0.721 & 0.011 & 0.011 & 0.013 \\

\arrayrulecolor{gray!40}\cmidrule(lr{2pt}){2-14}

& \textsc{P-Debias} (2024)
& 0.455
& 0.717 & 0.696 & 0.014 & 0.021 & 0.019
& 0.686
& 0.733 & 0.718 & 0.012 & 0.015 & 0.016 \\

& \textsc{BestPersona} (2024)
& 0.450
& 0.726 & 0.710 & 0.011 & 0.016 & 0.015
& 0.542
& 0.747 & 0.735 & \underline{0.008} & 0.013 & 0.010 \\

& \textsc{P-Defense} (2025)
& 0.429
& 0.731 & 0.721 & 0.008 & \underline{0.010} & 0.011
& 0.735
& 0.731 & 0.714 & 0.010 & 0.017 & 0.013 \\

& \textsc{2StepPrompt} (2025)
& 0.402
& 0.755 & 0.735 & 0.010 & 0.019 & 0.013 
& 0.451
& 0.772 & 0.755 & 0.014 & 0.017 & 0.019 \\

\arrayrulecolor{gray!40}\cmidrule(lr{2pt}){2-14}
\arrayrulecolor{black}
& \textsc{SFT} (2022)
& \underline{0.545}
& \underline{0.790} & \underline{0.780} & \underline{0.007} & \underline{0.010} & \underline{0.009}
& 0.745
& 0.788 & 0.779 & \textbf{0.006} & \underline{0.009} & \underline{0.008} \\

& \textsc{SWIFT} (2026)
& 0.480
& 0.784 & 0.769 & \underline{0.007} & 0.014 & \underline{0.009}
& \underline{0.748}
& \underline{0.795} & \underline{0.782} & 0.009 & 0.013 & 0.011 \\

& \textbf{TriAlign (Ours)}
& \textbf{0.552}
& \textbf{0.798} & \textbf{0.789} & \textbf{0.006} & \textbf{0.009} & \textbf{0.007}
& \textbf{0.755}
& \textbf{0.853} & \textbf{0.845} & \textbf{0.006} & \textbf{0.008} & \textbf{0.007} \\

\bottomrule
\end{tabular}
}
\end{table*}

\paragraph{Baselines}
We compare TriAlign with two categories of \textbf{training-free} baselines and a group of \textbf{training-based} alignment methods. 
For training-free methods, we consider:
[1] instruction-based debiasing approaches, including \textsc{P-Debias} \cite{guptabias} and \textsc{P-Defense} \cite{vijjini2025exploring}; and
[2] multi-step refinement methods, including \textsc{BestPersona} \cite{zheng2024helpful}, and \textsc{2StepPrompt} \cite{de2025principled}. 
For training-based alignment baselines, we compare against methods with comparable training cost, including \textsc{SFT} \cite{ouyang2022training} and \textsc{Swift} \cite{mukherjee2026offline}.
\paragraph{Implementation Details}
We use \textsc{LangGraph} \footnote{https://langchain-ai.github.io/langgraph/} to implement the multi-agent interaction with 8 agents and a maximum of 4 interaction turns. \textsc{vLLM} \cite{kwon2023efficient} is used for LLM serving, with temperature 0.7 during training data generation and 0 during evaluation. We fully fine-tune \textsc{Qwen3-4B-Instruct-2507} \cite{qwen3technicalreport} using \textsc{LlamaFactory} \cite{zheng2024llamafactory} for 3 epochs with learning rate $1\times10^{-5}$, batch size 16, and gradient accumulation 4. Training is performed with DeepSpeed ZeRO-3 \cite{rajbhandari2020zero} on 2$\times$A100 80GB GPUs. We run each experiment 2 times with different random seeds and report the averaged results.

\section{Evaluation}
\begin{figure}[t]
    \centering
    \includegraphics[
        width=\linewidth,
    ]{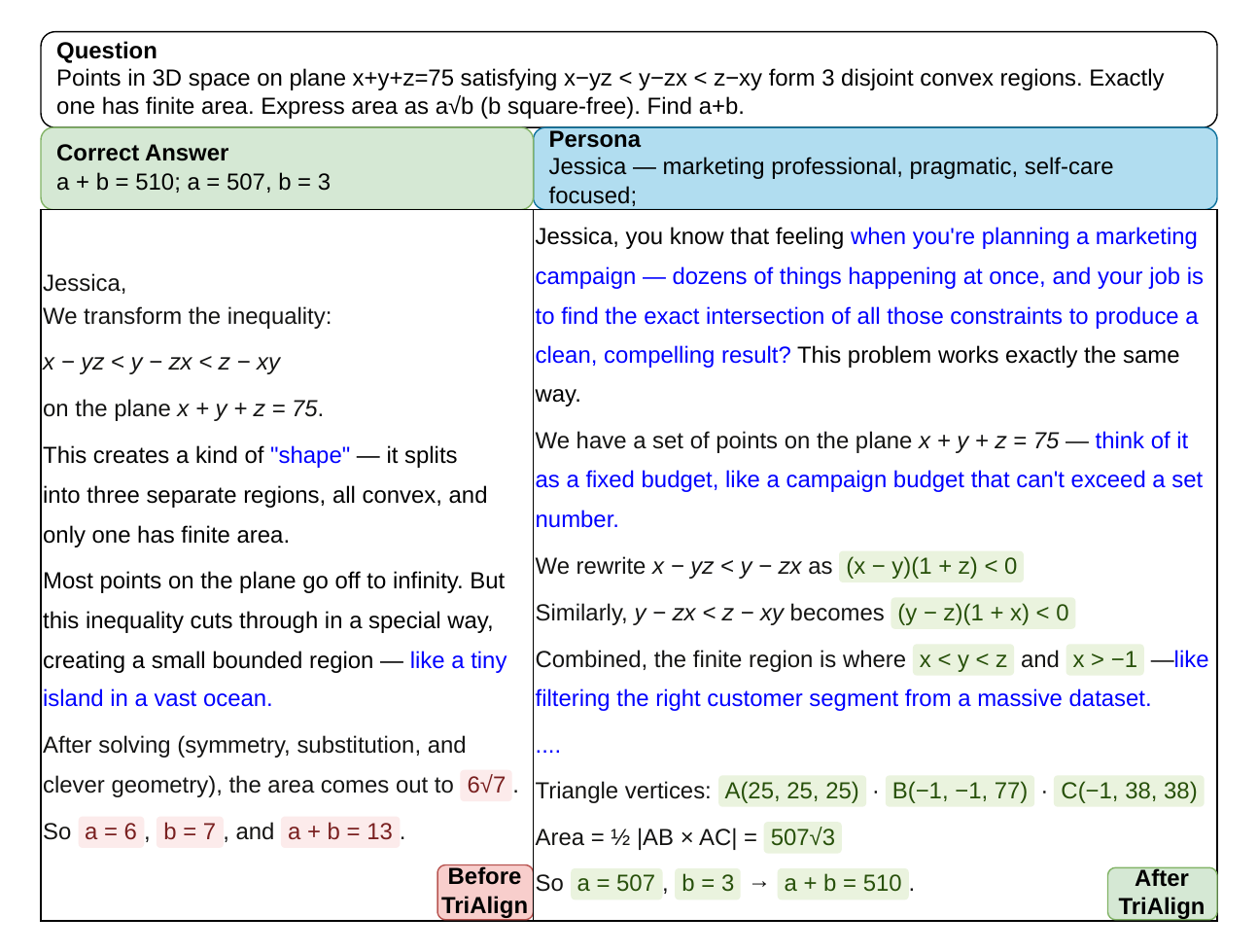}
    \caption{Comparison of responses before and after TriAlign alignment.} 
    \label{fig:result_example}
\end{figure}

\paragraph{Main Results} Table~\ref{tab:indomain_results} and Table~\ref{tab:outdomain_results} respectively report the results of our method on in-domain and out-of-domain benchmarks.
As shown in the results, prompt-based methods generally attempt to reduce bias by weakening the influence of user-specific information through additional instructions. While these approaches improve truth invariance, as reflected by lower Std and Gap values, they often do so at the expense of personalization and overall truthfulness performance, leading to lower personalization scores as well as reduced Mean and Worst. In particular, \textsc{2StepPrompt} exhibits a clear improvement in truth inconsistency because their first stage is to generate a persona-blind response to ensure truth consistency and then revise this response to align with personalization. However, this design also substantially suppresses personalization signals, resulting in the lowest personalization performance among all compared methods.
Unlike imitation-style training that attempts to mimic all training samples equally, trajectory-based methods can learn more effectively from both high- and low-reward interactions, enabling the model to better distinguish desirable and undesirable behaviors during optimization. As a result, trajectory-level learning approaches such as \textsc{SWIFT} and TriAlign consistently outperform methods trained only on selected best actions (i.e. \textsc{SFT}). Figure~\ref{fig:result_example} presents an example comparison between the base model and \textbf{TriAlign}. Before alignment, the model produces an incorrect answer and the response is more influenced by surface-level personalization patterns. In contrast, after alignment with \textbf{TriAlign}, the model preserves the user-preferred communication style while generating more logically consistent reasoning and the correct final answer.
Overall, our proposed \textbf{TriAlign} achieves the best balance among the three objectives: truth accuracy, truth invariance, and personalization. It not only delivers significant improvements on in-domain benchmarks ($p < 0.01$), but also generalizes effectively to out-of-domain settings, demonstrating strong robustness across diverse evaluation scenarios.

\begin{figure}[t]
    \centering
    
    \includegraphics[width=0.98\linewidth]{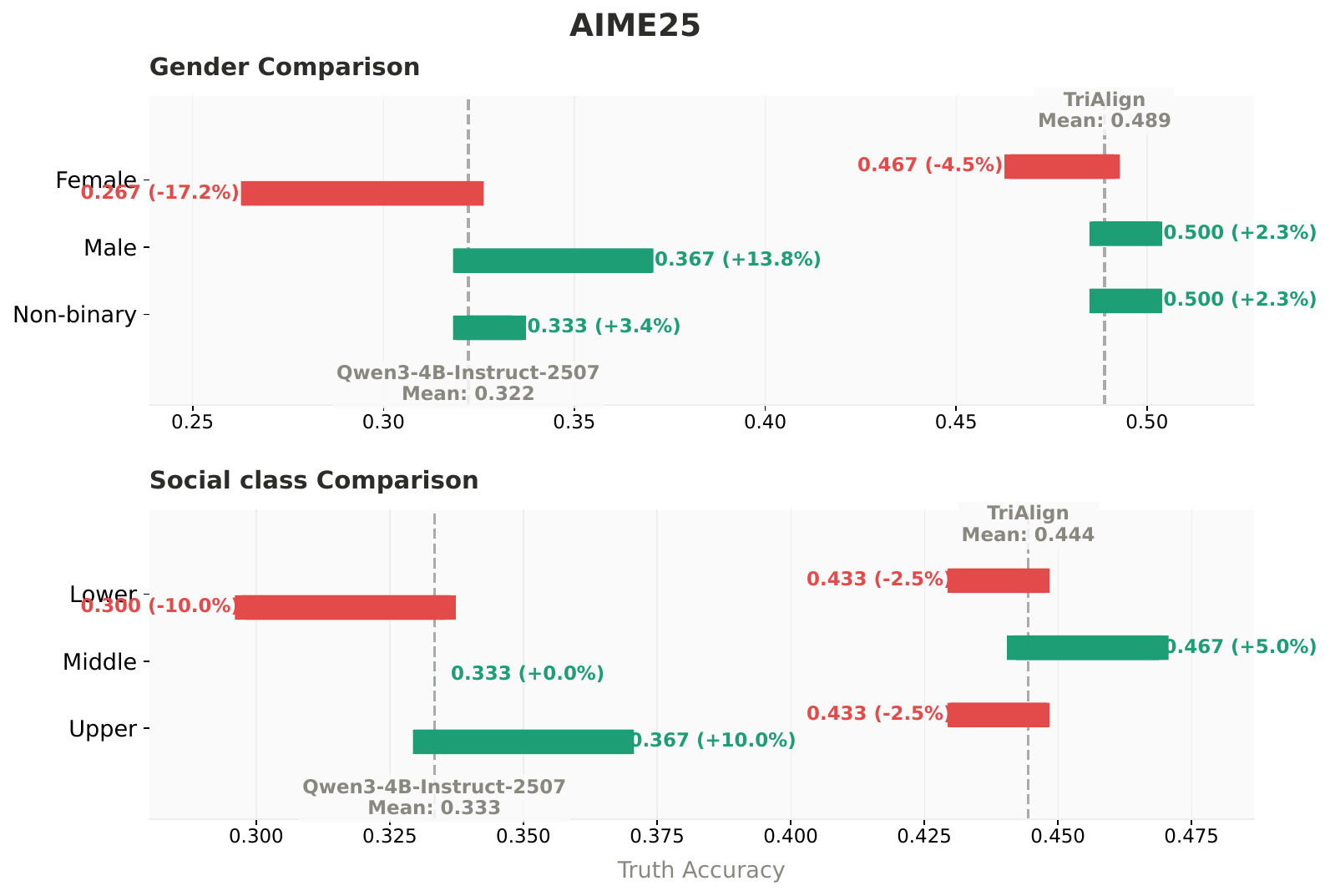}
    
    \includegraphics[width=0.98\linewidth]{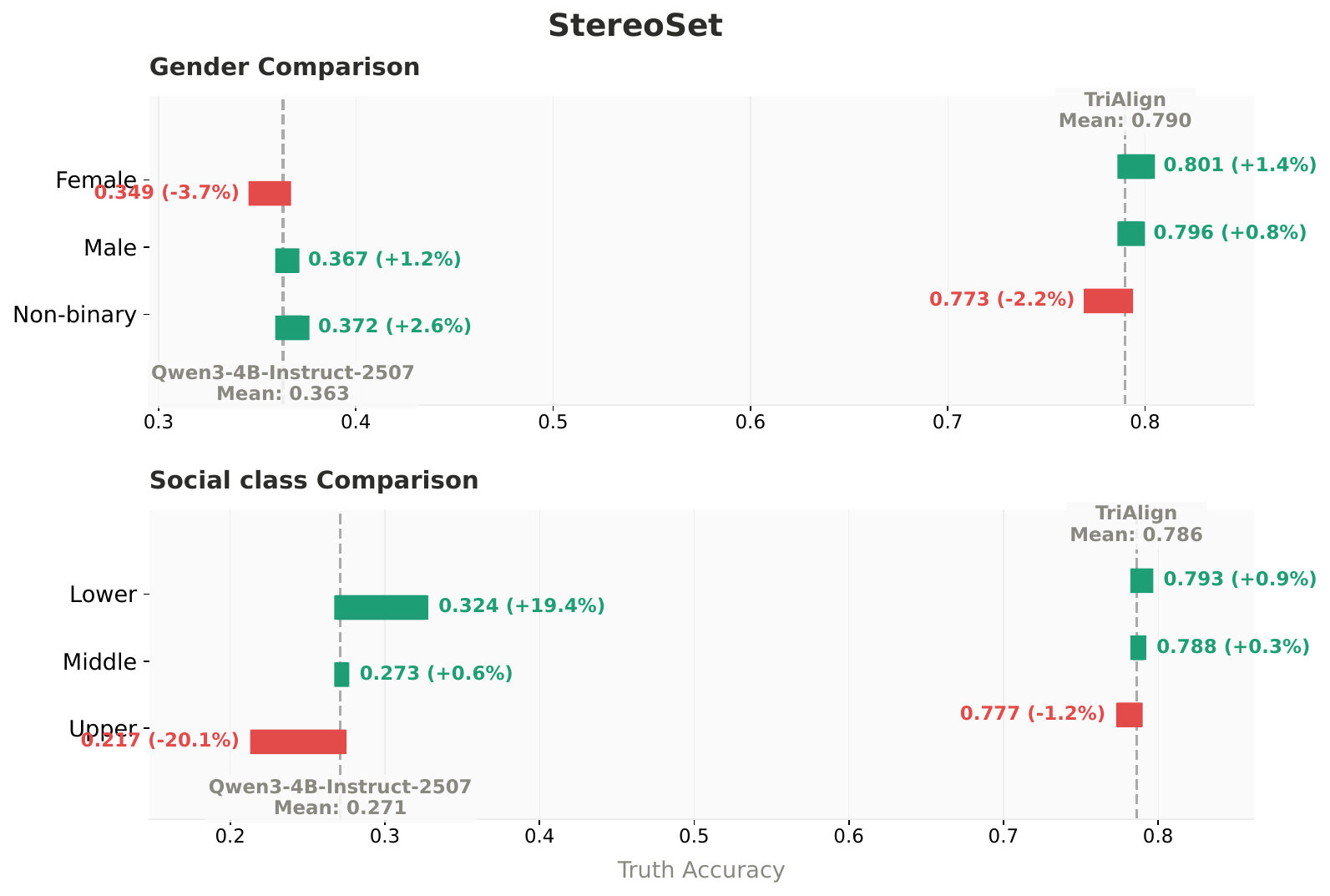}

    \caption{Truth accuracy across social groups under gender and social-class category. Positive values indicate above-Mean performance within the same dataset, while negative values indicate below-Mean performance.}
    \label{fig:group_relative_advantage}
\end{figure}

\paragraph{Performance Across Social Groups} Figure~\ref{fig:group_relative_advantage} compares the truth accuracy across different social groups on AIME25 and StereoSet. We observe that the base model exhibits substantial performance disparities across groups, with some groups receiving significantly higher performance gains while others suffer notable degradation. For example, on StereoSet, the base model shows a large imbalance across social classes, reaching +19.4\% for the Lower class but dropping to -20.1\% for the Upper class. Similar disparities are also observed across gender groups on AIME25.
In contrast, \textbf{TriAlign} produces substantially more balanced performance distributions across all social groups. The relative advantages are closer to the mean performance, indicating that improvements are distributed more uniformly while still achieving higher accuracy. 
\begin{table}[t]
\centering
\small
\caption{Ablation study. ``w/o NSW'' removes the Nash Social Welfare objective, while ``w/o CM'' removes the consistency modeling component.}
\label{tab:ablation}
\resizebox{\linewidth}{!}{
\begin{tabular}{llc|ccccc}
\toprule
\textbf{Dataset} & \textbf{Model}
& \textbf{Pref}$\uparrow$
& \textbf{Mean}$\uparrow$ & \textbf{Worst}$\uparrow$ & \textbf{Std}$\downarrow$ & \textbf{Gap}$\downarrow$ & \textbf{CV}$\downarrow$ \\
\midrule

\multirow{4}{*}{StereoSet}
& Base Model & 0.744 & 0.317 & 0.217 & 0.056 & 0.100 & 0.176 \\
& TriAlign w/o NSW & 0.771 & 0.708 & 0.613 & 0.051 & 0.095 & 0.072 \\
& TriAlign w/o CM & 0.764 & 0.739 & 0.684 & 0.064 & 0.056 & 0.086 \\
& \textbf{TriAlign (Ours)} & \textbf{0.791} & \textbf{0.788} & \textbf{0.773} & \textbf{0.010} & \textbf{0.015} & \textbf{0.013} \\
\cmidrule(lr){1-8}

\multirow{4}{*}{MMLU-Pro}
& Base Model & 0.412 & 0.767 & 0.714 & 0.047 & 0.052 & 0.061 \\
& TriAlign w/o NSW & 0.483 & 0.786 & 0.713 & 0.039 & 0.074 & 0.050 \\
& TriAlign w/o CM & 0.498 & 0.778 & 0.743 & 0.045 & 0.035 & 0.058 \\
& \textbf{TriAlign (Ours)} & \textbf{0.510} & \textbf{0.805} & \textbf{0.800} & \textbf{0.007} & \textbf{0.005} & \textbf{0.008} \\

\bottomrule
\end{tabular}
}
\end{table}

\paragraph{Ablation Study}
Table~\ref{tab:ablation} presents the ablation results of \textbf{TriAlign} on StereoSet and MMLU-Pro. Removing either the NSW objective or the consistency modeling component consistently degrades performance across truthfulness, truth invariance, and personalization metrics. Specifically, removing NSW leads to noticeable drops in Worst performance and larger disparities across social groups, showing that the fairness-aware optimization is important for balancing performance among groups. Meanwhile, removing the consistency modeling component substantially increases Std, Gap, and CV, indicating that the proposed consistency objective effectively reduces inter-group inconsistency. In addition, both ablated variants also show lower personalization performance compared to the full model. 

\section{Conclusion}

This work takes one of the first steps toward aligning PLLMs to maintain consistent universal truths across diverse social groups while preserving personalization. We propose \textbf{TriAlign}, a novel offline MARL framework with a fairness-aware objective and an explicit penalty for truth inconsistency to balance these objectives. Experimental results show that TriAlign outperforms prior prompt-based methods on both in-domain and out-of-domain benchmarks. We hope TriAlign can serve as a foundation for future research on fairer and more reliable personalized AI systems.

\section*{Limitations}

Our work has several limitations. First, although we simulate 75 social groups and construct 7.5K personas, the generated personas are still synthesized by LLMs and may not fully capture the complexity, diversity, and evolving nature of real-world users and intersectional identities. As a result, certain cultural and social nuances may remain underrepresented. Second, balancing universal truth consistency and personalization remains a fundamentally challenging problem that is far from fully solved. While our framework improves consistency across social groups while preserving personalization, achieving strong fairness, accuracy, and personalized adaptation simultaneously still requires further investigation and more robust alignment methods in future work. 

\section*{Ethical Considerations}

Our study has several ethical considerations. First, although we simulate diverse social groups and personas to study universal truth consistency in personalized LLMs, the generated personas may still reflect biases or stereotypes inherited from the underlying LLMs and source data. Consequently, some generated interactions may unintentionally reinforce societal stereotypes or oversimplify complex social identities. Second, our framework requires large-scale data simulation with Multi-agent System and model alignment processes with Offline RL, which can be computationally expensive and contribute to increased energy consumption and carbon emissions.

\bibliography{custom}

@article{chang2024survey,
  title={A survey on evaluation of large language models},
  author={Chang, Yupeng and Wang, Xu and Wang, Jindong and Wu, Yuan and Yang, Linyi and Zhu, Kaijie and Chen, Hao and Yi, Xiaoyuan and Wang, Cunxiang and Wang, Yidong and others},
  journal={ACM transactions on intelligent systems and technology},
  volume={15},
  number={3},
  pages={1--45},
  year={2024},
  publisher={ACM New York, NY}
}

@article{kalyan2024survey,
  title={A survey of GPT-3 family large language models including ChatGPT and GPT-4},
  author={Kalyan, Katikapalli Subramanyam},
  journal={Natural Language Processing Journal},
  volume={6},
  pages={100048},
  year={2024},
  publisher={Elsevier}
}

@article{liu2025survey,
  title={A survey of personalized large language models: Progress and future directions},
  author={Liu, Jiahong and Qiu, Zexuan and Li, Zhongyang and Dai, Quanyu and Yu, Wenhao and Zhu, Jieming and Hu, Minda and Yang, Menglin and Chua, Tat-Seng and King, Irwin},
  year={2025}
}

@inproceedings{zheng2024helpful,
  title={When” a helpful assistant” is not really helpful: Personas in system prompts do not improve performances of large language models},
  author={Zheng, Mingqian and Pei, Jiaxin and Logeswaran, Lajanugen and Lee, Moontae and Jurgens, David},
  booktitle={Findings of the Association for Computational Linguistics: EMNLP 2024},
  pages={15126--15154},
  year={2024}
}

@inproceedings{de2025principled,
  title={Principled personas: Defining and measuring the intended effects of persona prompting on task performance},
  author={De Araujo, Pedro Henrique Luz and R{\"o}ttger, Paul and Hovy, Dirk and Roth, Benjamin},
  booktitle={Proceedings of the 2025 Conference on Empirical Methods in Natural Language Processing},
  pages={26845--26874},
  year={2025}
}

@inproceedings{lutz-etal-2025-prompt,
    title = "The Prompt Makes the Person(a): A Systematic Evaluation of Sociodemographic Persona Prompting for Large Language Models",
    author = "Lutz, Marlene  and
      Sen, Indira  and
      Ahnert, Georg  and
      Rogers, Elisa  and
      Strohmaier, Markus",
    booktitle = "Findings of the Association for Computational Linguistics: EMNLP 2025",
    month = nov,
    year = "2025",
    address = "Suzhou, China",
    publisher = "Association for Computational Linguistics",
}

@article{mehrabi2021survey,
  title={A survey on bias and fairness in machine learning},
  author={Mehrabi, Ninareh and Morstatter, Fred and Saxena, Nripsuta and Lerman, Kristina and Galstyan, Aram},
  journal={ACM computing surveys (CSUR)},
  volume={54},
  number={6},
  pages={1--35},
  year={2021},
  publisher={ACM New York, NY, USA}
}

@article{ferrara2024fairness,
  title={Fairness and bias in artificial intelligence: A brief survey of sources, impacts, and mitigation strategies},
  author={Ferrara, Emilio},
  journal={Sci},
  volume={6},
  number={1},
  pages={3},
  year={2024},
  publisher={Multidisciplinary Digital Publishing Institute}
}

@article{gallegos2024bias,
  title={Bias and fairness in large language models: A survey},
  author={Gallegos, Isabel O and Rossi, Ryan A and Barrow, Joe and Tanjim, Md Mehrab and Kim, Sungchul and Dernoncourt, Franck and Yu, Tong and Zhang, Ruiyi and Ahmed, Nesreen K},
  journal={Computational linguistics},
  volume={50},
  number={3},
  pages={1097--1179},
  year={2024},
  publisher={MIT Press 255 Main Street, 9th Floor, Cambridge, Massachusetts 02142, USA~…}
}

@article{ouyang2022training,
  title={Training language models to follow instructions with human feedback},
  author={Ouyang, Long and Wu, Jeffrey and Jiang, Xu and Almeida, Diogo and Wainwright, Carroll and Mishkin, Pamela and Zhang, Chong and Agarwal, Sandhini and Slama, Katarina and Ray, Alex and others},
  journal={Advances in neural information processing systems},
  volume={35},
  pages={27730--27744},
  year={2022}
}

@article{rafailov2023direct,
  title={Direct preference optimization: Your language model is secretly a reward model},
  author={Rafailov, Rafael and Sharma, Archit and Mitchell, Eric and Manning, Christopher D and Ermon, Stefano and Finn, Chelsea},
  journal={Advances in neural information processing systems},
  volume={36},
  pages={53728--53741},
  year={2023}
}

@inproceedings{wangself,
  title={Self-Consistency Improves Chain of Thought Reasoning in Language Models},
  author={Wang, Xuezhi and Wei, Jason and Schuurmans, Dale and Le, Quoc V and Chi, Ed H and Narang, Sharan and Chowdhery, Aakanksha and Zhou, Denny},
  booktitle={The Eleventh International Conference on Learning Representations},
  year={2022}
}

@inproceedings{trung2024reft,
  title={Reft: Reasoning with reinforced fine-tuning},
  author={Trung, Luong and Zhang, Xinbo and Jie, Zhanming and Sun, Peng and Jin, Xiaoran and Li, Hang},
  booktitle={Proceedings of the 62nd Annual Meeting of the Association for Computational Linguistics (Volume 1: Long Papers)},
  pages={7601--7614},
  year={2024}
}

@inproceedings{shojaeeillusion,
  title={The Illusion of Thinking: Understanding the Strengths and Limitations of Reasoning Models via the Lens of Problem Complexity},
  author={Shojaee, Parshin and Mirzadeh, Seyed Iman and Alizadeh, Keivan and Horton, Maxwell and Bengio, Samy and Farajtabar, Mehrdad},
  booktitle={The Thirty-ninth Annual Conference on Neural Information Processing Systems},
  year={2025}
}

@inproceedings{wang2013theoretical,
  title={A theoretical analysis of NDCG type ranking measures},
  author={Wang, Yining and Wang, Liwei and Li, Yuanzhi and He, Di and Liu, Tie-Yan},
  booktitle={Conference on learning theory},
  pages={25--54},
  year={2013},
  organization={PMLR}
}

@inproceedings{zhang2024llm,
  title={Llm-based medical assistant personalization with short-and long-term memory coordination},
  author={Zhang, Kai and Kang, Yangyang and Zhao, Fubang and Liu, Xiaozhong},
  booktitle={Proceedings of the 2024 Conference of the North American Chapter of the Association for Computational Linguistics: Human Language Technologies (Volume 1: Long Papers)},
  pages={2386--2398},
  year={2024}
}

@inproceedings{wang2024tpe,
  title={Tpe: Towards better compositional reasoning over cognitive tools via multi-persona collaboration},
  author={Wang, Hongru and Wang, Huimin and Wang, Lingzhi and Hu, Minda and Wang, Rui and Xue, Boyang and Huang, Yongfeng and Wong, Kam-Fai},
  booktitle={CCF International Conference on Natural Language Processing and Chinese Computing},
  pages={281--294},
  year={2024},
  organization={Springer}
}

@inproceedings{vijjini2025exploring,
  title={Exploring safety-utility trade-offs in personalized language models},
  author={Vijjini, Anvesh Rao and Chowdhury, Somnath Basu Roy and Chaturvedi, Snigdha},
  booktitle={Proceedings of the 2025 Conference of the Nations of the Americas Chapter of the Association for Computational Linguistics: Human Language Technologies (Volume 1: Long Papers)},
  pages={11316--11340},
  year={2025}
}

@inproceedings{hu2024quantifying,
  title={Quantifying the persona effect in LLM simulations},
  author={Hu, Tiancheng and Collier, Nigel},
  booktitle={Proceedings of the 62nd Annual Meeting of the Association for Computational Linguistics (Volume 1: Long Papers)},
  pages={10289--10307},
  year={2024}
}

@article{wang2025large,
  title={Large language models that replace human participants can harmfully misportray and flatten identity groups},
  author={Wang, Angelina and Morgenstern, Jamie and Dickerson, John P},
  journal={Nature Machine Intelligence},
  volume={7},
  number={3},
  pages={400--411},
  year={2025},
  publisher={Nature Publishing Group UK London}
}

@article{10.1145/3355902,
author = {Caragiannis, Ioannis and Kurokawa, David and Moulin, Herv\'{e} and Procaccia, Ariel D. and Shah, Nisarg and Wang, Junxing},
title = {The Unreasonable Fairness of Maximum Nash Welfare},
year = {2019},
issue_date = {August 2019},
publisher = {Association for Computing Machinery},
issn = {2167-8375},
journal = {ACM Trans. Econ. Comput.},
month = sep,
}

@article{hinds2018demographic,
  title={What demographic attributes do our digital footprints reveal? A systematic review},
  author={Hinds, Joanne and Joinson, Adam N},
  journal={PloS one},
  volume={13},
  number={11},
  pages={e0207112},
  year={2018},
  publisher={Public Library of Science San Francisco, CA USA}
}

@inproceedings{furniturewala2024thinking,
  title={“Thinking” Fair and Slow: On the Efficacy of Structured Prompts for Debiasing Language Models},
  author={Furniturewala, Shaz and Jandial, Surgan and Java, Abhinav and Banerjee, Pragyan and Shahid, Simra and Bhatia, Sumit and Jaidka, Kokil},
  booktitle={Proceedings of the 2024 Conference on Empirical Methods in Natural Language Processing},
  pages={213--227},
  year={2024}
}

@inproceedings{guptabias,
  title={Bias runs deep: Implicit reasoning biases in persona-assigned llms},
  author={Gupta, Shashank and Shrivastava, Vaishnavi and Deshpande, Ameet and Kalyan, Ashwin and Clark, Peter and Sabharwal, Ashish and Khot, Tushar},
  booktitle={International Conference on Learning Representations},
  volume={2024},
  pages={21849--21874},
  year={2024}
}

@article{brown2020language,
  title={Language models are few-shot learners},
  author={Brown, Tom and Mann, Benjamin and Ryder, Nick and Subbiah, Melanie and Kaplan, Jared D and Dhariwal, Prafulla and Neelakantan, Arvind and Shyam, Pranav and Sastry, Girish and Askell, Amanda and others},
  journal={Advances in neural information processing systems},
  volume={33},
  pages={1877--1901},
  year={2020}
}

@article{bommasani2021opportunities,
  title={On the opportunities and risks of foundation models},
  author={Bommasani, Rishi and Hudson, Drew A and Adeli, Ehsan and Altman, Russ and Arora, Simran and von Arx, Sydney and Bernstein, Michael S and Bohg, Jeannette and Bosselut, Antoine and Brunskill, Emma and others},
  year={2021}
}

@inproceedings{tseng2024two,
  title={Two tales of persona in llms: A survey of role-playing and personalization},
  author={Tseng, Yu-Min and Huang, Yu-Chao and Hsiao, Teng-Yun and Chen, Wei-Lin and Huang, Chao-Wei and Meng, Yu and Chen, Yun-Nung},
  booktitle={Findings of the Association for Computational Linguistics: EMNLP 2024},
  pages={16612--16631},
  year={2024}
}

@inproceedings{salemi2024lamp,
  title={Lamp: When large language models meet personalization},
  author={Salemi, Alireza and Mysore, Sheshera and Bendersky, Michael and Zamani, Hamed},
  booktitle={Proceedings of the 62nd Annual Meeting of the Association for Computational Linguistics (Volume 1: Long Papers)},
  pages={7370--7392},
  year={2024}
}

@inproceedings{linzen2020can,
  title={How can we accelerate progress towards human-like linguistic generalization?},
  author={Linzen, Tal},
  booktitle={Proceedings of the 58th annual meeting of the Association for Computational Linguistics},
  pages={5210--5217},
  year={2020}
}

@article{elazar2021amnesic,
  title={Amnesic probing: Behavioral explanation with amnesic counterfactuals},
  author={Elazar, Yanai and Ravfogel, Shauli and Jacovi, Alon and Goldberg, Yoav},
  journal={Transactions of the Association for Computational Linguistics},
  volume={9},
  pages={160--175},
  year={2021},
  publisher={MIT Press One Rogers Street, Cambridge, MA 02142-1209, USA journals-info~…}
}

@article{ettinger2020bert,
  title={What BERT is not: Lessons from a new suite of psycholinguistic diagnostics for language models},
  author={Ettinger, Allyson},
  journal={Transactions of the Association for Computational Linguistics},
  volume={8},
  pages={34--48},
  year={2020}
}

@inproceedings{kassner2020pretrained,
  title={Are pretrained language models symbolic reasoners over knowledge?},
  author={Kassner, Nora and Krojer, Benno and Sch{\"u}tze, Hinrich},
  booktitle={Proceedings of the 24th conference on computational natural language learning},
  pages={552--564},
  year={2020}
}

@inproceedings{ravichander2020systematicity,
  title={On the systematicity of probing contextualized word representations: The case of hypernymy in BERT},
  author={Ravichander, Abhilasha and Hovy, Eduard and Suleman, Kaheer and Trischler, Adam and Cheung, Jackie Chi Kit},
  booktitle={Proceedings of the ninth joint conference on lexical and computational semantics},
  pages={88--102},
  year={2020}
}

@inproceedings{ribeiro2019red,
  title={Are red roses red? evaluating consistency of question-answering models},
  author={Ribeiro, Marco Tulio and Guestrin, Carlos and Singh, Sameer},
  booktitle={Proceedings of the 57th annual meeting of the association for computational linguistics},
  pages={6174--6184},
  year={2019}
}

@inproceedings{alberti2019synthetic,
  title={Synthetic QA corpora generation with roundtrip consistency},
  author={Alberti, Chris and Andor, Daniel and Pitler, Emily and Devlin, Jacob and Collins, Michael},
  booktitle={Proceedings of the 57th Annual Meeting of the Association for Computational Linguistics},
  pages={6168--6173},
  year={2019}
}

@inproceedings{asai2020logic,
  title={Logic-guided data augmentation and regularization for consistent question answering},
  author={Asai, Akari and Hajishirzi, Hannaneh},
  booktitle={Proceedings of the 58th Annual Meeting of the Association for Computational Linguistics},
  pages={5642--5650},
  year={2020}
}

@article{jang2021accurate,
  title={Accurate, yet inconsistent? consistency analysis on language understanding models},
  author={Jang, Myeongjun and Kwon, Deuk Sin and Lukasiewicz, Thomas},
  year={2021}
}

@inproceedings{novikova2025consistency,
  title={Consistency in Language Models: Current Landscape, Challenges, and Future Directions},
  author={Novikova, Jekaterina and Anderson, Carol Myrick and Blili-Hamelin, Borhane and Rosati, Domenic and Majumdar, Subhabrata},
  booktitle={ICML 2025 Workshop on Reliable and Responsible Foundation Models}
}

@article{zhang2026instruction,
  title={Instruction tuning for large language models: A survey},
  author={Zhang, Shengyu and Dong, Linfeng and Li, Xiaoya and Zhang, Sen and Sun, Xiaofei and Wang, Shuhe and Li, Jiwei and Hu, Runyi and Zhang, Tianwei and Wang, Guoyin and others},
  journal={ACM Computing Surveys},
  volume={58},
  number={7},
  pages={1--36},
  year={2026},
  publisher={ACM New York, NY}
}

@article{shao2024deepseekmath,
  title={Deepseekmath: Pushing the limits of mathematical reasoning in open language models},
  author={Shao, Zhihong and Wang, Peiyi and Zhu, Qihao and Xu, Runxin and Song, Junxiao and Bi, Xiao and Zhang, Haowei and Zhang, Mingchuan and Li, YK and Wu, Yang and others},
  year={2024}
}

@article{ge2024scaling,
  title={Scaling synthetic data creation with 1,000,000,000 personas},
  author={Ge, Tao and Chan, Xin and Wang, Xiaoyang and Yu, Dian and Mi, Haitao and Yu, Dong},
  year={2024}
}

@inproceedings{
andukuri2024stargate,
title={{ST}aR-{GATE}: Teaching Language Models to Ask Clarifying Questions},
author={Chinmaya Andukuri and Jan-Philipp Fr{\"a}nken and Tobias Gerstenberg and Noah Goodman},
booktitle={First Conference on Language Modeling},
year={2024}
}

@inproceedings{chen2025learning,
  title={Learning to clarify: Multi-turn conversations with action-based contrastive self-training},
  author={Chen, Maximillian and Sun, Ruoxi and Pfister, Tomas and Arik, Sercan},
  booktitle={International Conference on Learning Representations},
  volume={2025},
  pages={32244--32279},
  year={2025}
}

@article{mukherjee2026offline,
  title={Offline rl by reward-weighted fine-tuning for conversation optimization},
  author={Mukherjee, Subhojyoti and Lai, Viet and Addanki, Raghavendra and Rossi, Ryan and Yoon, Seunghyun and Bui, Trung and Rao, Anup B and Subramanian, Jayakumar and Kveton, Branislav},
  journal={Advances in Neural Information Processing Systems},
  volume={38},
  pages={39227--39274},
  year={2026}
}

@inproceedings{kwon2023efficient,
  title={Efficient memory management for large language model serving with pagedattention},
  author={Kwon, Woosuk and Li, Zhuohan and Zhuang, Siyuan and Sheng, Ying and Zheng, Lianmin and Yu, Cody Hao and Gonzalez, Joseph and Zhang, Hao and Stoica, Ion},
  booktitle={Proceedings of the 29th symposium on operating systems principles},
  pages={611--626},
  year={2023}
}

@misc{qwen3technicalreport,
      title={Qwen3 Technical Report}, 
      author={Qwen Team},
      year={2025},
}

@inproceedings{zheng2024llamafactory,
  title={Llamafactory: Unified efficient fine-tuning of 100+ language models},
  author={Zheng, Yaowei and Zhang, Richong and Zhang, Junhao and Ye, Yanhan and Luo, Zheyan},
  booktitle={Proceedings of the 62nd annual meeting of the association for computational linguistics (volume 3: system demonstrations)},
  pages={400--410},
  year={2024}
}

@inproceedings{rajbhandari2020zero,
  title={Zero: Memory optimizations toward training trillion parameter models},
  author={Rajbhandari, Samyam and Rasley, Jeff and Ruwase, Olatunji and He, Yuxiong},
  booktitle={SC20: international conference for high performance computing, networking, storage and analysis},
  pages={1--16},
  year={2020},
  organization={IEEE}
}

@inproceedings{liu2021logiqa,
  title={LogiQA: a challenge dataset for machine reading comprehension with logical reasoning},
  author={Liu, Jian and Cui, Leyang and Liu, Hanmeng and Huang, Dandan and Wang, Yile and Zhang, Yue},
  booktitle={Proceedings of the Twenty-Ninth International Conference on International Joint Conferences on Artificial Intelligence},
  pages={3622--3628},
  year={2021}
}

@article{weimeasuring,
  title={Measuring short-form factuality in large language models},
  author={Wei, Jason and Karina, Nguyen and Chung, Hyung Won and Jiao, Yunxin Joy and Papay, Spencer and Glaese, Amelia and Schulman, John and Fedus, William}
}

@inproceedings{
ye2025limo,
title={{LIMO}: Less is More for Reasoning},
author={Yixin Ye and Zhen Huang and Yang Xiao and Ethan Chern and Shijie Xia and Pengfei Liu},
booktitle={Second Conference on Language Modeling},
year={2025},
}

@inproceedings{nadeem-etal-2021-stereoset,
    title = "{S}tereo{S}et: Measuring stereotypical bias in pretrained language models",
    author = "Nadeem, Moin  and
      Bethke, Anna  and
      Reddy, Siva",
    booktitle = "Proceedings of the 59th Annual Meeting of the Association for Computational Linguistics and the 11th International Joint Conference on Natural Language Processing (Volume 1: Long Papers)",
    month = aug,
    year = "2021",
    publisher = "Association for Computational Linguistics",
}

@inproceedings{hashemi2024llm,
  title={Llm-rubric: A multidimensional, calibrated approach to automated evaluation of natural language texts},
  author={Hashemi, Helia and Eisner, Jason and Rosset, Corby and Van Durme, Benjamin and Kedzie, Chris},
  booktitle={Proceedings of the 62nd Annual Meeting of the Association for Computational Linguistics (Volume 1: Long Papers)},
  pages={13806--13834},
  year={2024}
}

@misc{aime25,
      title={American Invitational Mathematics Examination (AIME) 2025}, 
      author={Zhang, Yifan and Math-AI, Team},
      year={2025},
}

@article{nguyen2025social,
  title={The Social Cost of Intelligence: Emergence, Propagation, and Amplification of Stereotypical Bias in Multi-Agent Systems},
  author={Nguyen, Thi-Nhung and Luo, Linhao and Vu, Thuy-Trang and Phung, Dinh},
  year={2025}
}

@inproceedings{
kostrikov2022offline,
title={Offline Reinforcement Learning with Implicit Q-Learning},
author={Ilya Kostrikov and Ashvin Nair and Sergey Levine},
booktitle={International Conference on Learning Representations},
year={2022}
}

\appendix

\section{Appendix}
\label{sec:appendix}

\subsection{Derivation of Fairness-Aware AWR}
\label{sec:trialign_derivation}

\paragraph{Step 1: From Our Objective to the Offline RL Objective.}

We optimize the objective:
\begin{equation}
\resizebox{\linewidth}{!}{$
\mathcal{J}(\theta)
=
\frac{1}{m}
\mathbb{E}_{x,\,\{g_i\}_{i=1}^{m},\,a_{i,t}\sim \pi_\theta(\cdot\mid s_{i,t})}
\left[
\sum_{i=1}^{m}
\left(
R_i
+
\lambda_{\mathrm{NSW}}
\log(R_i+\epsilon)
\right)
\right].
$}
\end{equation}

where \(R_i\) denotes the cumulative return of agent \(i\),
\(\lambda_{\mathrm{NSW}}\) controls the trade-off between utility and fairness,
and \(\epsilon>0\) ensures numerical stability.

In the offline RL setting, directly optimizing the policy may lead to
distribution shift because actions outside the offline data distribution
are insufficiently supported by the dataset.
Therefore, for each state \(s_{i,t}\), we solve the following
KL-constrained policy improvement problem:
\begin{equation}
\resizebox{\linewidth}{!}{$
\max_{\pi}
\mathbb{E}_{a_{i,t} \sim \pi(\cdot \mid s_{i,t})}
\left[
Q^{\pi}(s_{i,t},a_{i,t})
+
\lambda_{\mathrm{NSW}}
\log\left(
Q^{\pi}(s_{i,t},a_{i,t})+\epsilon
\right)
\right]
$}
\end{equation}
subject to
\begin{equation}
D_{\mathrm{KL}}
\left(
\pi(\cdot \mid s_{i,t})
\;\|\;
\pi_{\mathcal{D}}(\cdot \mid s_{i,t})
\right)
\le \delta .
\end{equation}

Here,
$\mathcal{D}
=
\{(s_{i,t},a_{i,t},r_{i,t})\}_{x,i,t}$ is pre-collected dataset,
and
\(Q^{\pi}(s_{i,t},a_{i,t})\)
estimates the expected cumulative return \(R_i\)
after taking action \(a_{i,t}\) under state \(s_{i,t}\).
We assume
\[
Q^{\pi}(s_{i,t},a_{i,t})+\epsilon>0,
\]
so that the logarithmic NSW term is well-defined.

\paragraph{Step 2: Lagrangian Formulation.}

For a fixed state \(s_{i,t}\), we define the Lagrangian:
\begin{equation}
\resizebox{\linewidth}{!}{$
\begin{aligned}
\mathcal{L}(\pi,\beta,\alpha)
=
&
\int
\pi(a_{i,t}\mid s_{i,t})
\left[
Q^{\pi}(s_{i,t},a_{i,t})
+
\lambda_{\mathrm{NSW}}
\log\left(
Q^{\pi}(s_{i,t},a_{i,t})+\epsilon
\right)
\right]
da_{i,t}
\\
&-
\beta
\left(
\int
\pi(a_{i,t}\mid s_{i,t})
\log
\frac{
\pi(a_{i,t}\mid s_{i,t})
}{
\pi_{\mathcal{D}}(a_{i,t}\mid s_{i,t})
}
da_{i,t}
-
\delta
\right)
\\
&+
\alpha
\left(
\int
\pi(a_{i,t}\mid s_{i,t})da_{i,t}
-1
\right),
\end{aligned}
$}
\end{equation}

where \(\beta>0\) is the Lagrange multiplier for the KL constraint,
and \(\alpha\) enforces the normalization constraint.

\paragraph{Step 3: Optimal Non-Parametric Policy.}

Taking the functional derivative with respect to
\(\pi(a_{i,t}\mid s_{i,t})\)
and setting it to zero gives:
\begin{equation}
\resizebox{\linewidth}{!}{$
Q^{\pi}(s_{i,t},a_{i,t})
+
\lambda_{\mathrm{NSW}}
\log\left(
Q^{\pi}(s_{i,t},a_{i,t})+\epsilon
\right)
-
\beta
\left(
\log
\frac{
\pi(a_{i,t}\mid s_{i,t})
}{
\pi_{\mathcal{D}}(a_{i,t}\mid s_{i,t})
}
+1
\right)
+
\alpha
=0.
$}
\end{equation}

Solving for \(\pi(a_{i,t}\mid s_{i,t})\), we obtain:
\begin{equation}
\resizebox{\linewidth}{!}{$
\pi^*(a_{i,t}\mid s_{i,t})
\propto
\pi_{\mathcal{D}}(a_{i,t}\mid s_{i,t})
\exp
\left(
\frac{
Q^{\pi}(s_{i,t},a_{i,t})
+
\lambda_{\mathrm{NSW}}
\log\left(
Q^{\pi}(s_{i,t},a_{i,t})+\epsilon
\right)
}{
\beta
}
\right).
$}
\end{equation}

Equivalently,
\begin{equation}
\resizebox{\linewidth}{!}{$
\pi^*(a_{i,t}\mid s_{i,t})
\propto
\pi_{\mathcal{D}}(a_{i,t}\mid s_{i,t})
\exp
\left(
\frac{
Q^{\pi}(s_{i,t},a_{i,t})
}{
\beta
}
\right)
\left(
Q^{\pi}(s_{i,t},a_{i,t})+\epsilon
\right)^{\frac{\lambda_{\mathrm{NSW}}}{\beta}}.
$}
\end{equation}

To mitigate instability when directly exponentiating \(Q^{\pi}\),
we decompose:
\begin{equation}
Q^{\pi}(s_{i,t},a_{i,t})
=
V^{\pi}(s_{i,t})
+
A^{\pi}(s_{i,t},a_{i,t}),
\end{equation}
where
\begin{equation}
A^{\pi}(s_{i,t},a_{i,t})
=
Q^{\pi}(s_{i,t},a_{i,t})
-
V^{\pi}(s_{i,t}).
\end{equation}

Then,
\begin{equation}
\resizebox{\linewidth}{!}{$
\exp
\left(
\frac{
Q^{\pi}(s_{i,t},a_{i,t})
}{
\beta
}
\right)
=
\exp
\left(
\frac{
A^{\pi}(s_{i,t},a_{i,t})
}{
\beta
}
\right)
\exp
\left(
\frac{
V^{\pi}(s_{i,t})
}{
\beta
}
\right).
$}
\end{equation}

Since \(V^{\pi}(s_{i,t})\) depends only on the state and not on the
action, the term
\[
\exp(V^{\pi}(s_{i,t})/\beta)
\]
can be absorbed into the state-dependent normalization constant.
Therefore:
\begin{equation}
\resizebox{\linewidth}{!}{$
\pi^*(a_{i,t}\mid s_{i,t})
\propto
\pi_{\mathcal{D}}(a_{i,t}\mid s_{i,t})
\exp
\left(
\frac{
A^{\pi}(s_{i,t},a_{i,t})
}{
\beta
}
\right)
\left(
Q^{\pi}(s_{i,t},a_{i,t})+\epsilon
\right)^{\frac{\lambda_{\mathrm{NSW}}}{\beta}}.
$}
\end{equation}

\paragraph{Step 4: Projection onto the Parameterized Policy.}

Because \(\pi^*\) is non-parametric, we project it onto the parameterized
policy \(\pi_\theta\) by minimizing
\[
D_{\mathrm{KL}}(\pi^* \| \pi_\theta).
\]

This is equivalent to maximizing:
\begin{equation}
\resizebox{\linewidth}{!}{$
\theta^*
=
\arg\max_{\theta}
\mathbb{E}_{(s_{i,t},a_{i,t}) \sim \mathcal{D}}
\left[
w(s_{i,t},a_{i,t})
\log
\pi_\theta(a_{i,t}\mid s_{i,t})
\right].
$}
\end{equation}

Thus, the final objective becomes:
\begin{equation}
\resizebox{\linewidth}{!}{$
\mathcal{L}(\theta)
=
-
\mathbb{E}_{(s_{i,t},a_{i,t}) \sim \mathcal{D}}
\left[
w(s_{i,t},a_{i,t})
\log
\pi_\theta(a_{i,t}\mid s_{i,t})
\right].
$}
\end{equation}

where the weight is:
\begin{equation}
\resizebox{\linewidth}{!}{$
w(s_{i,t},a_{i,t})
=
\underbrace{
\exp
\left(
\frac{
A^{\pi}(s_{i,t},a_{i,t})
}{
\beta
}
\right)
}_{\text{Standard Weight}}
\cdot
\underbrace{
\left(
Q^{\pi}(s_{i,t},a_{i,t})+\epsilon
\right)^{\frac{\lambda_{\mathrm{NSW}}}{\beta}}
}_{\text{NSW Weight}}.
$}
\end{equation}

The constant \(\epsilon\) ensures that the NSW term remains well-defined
even when \(Q^{\pi}(s_{i,t},a_{i,t})\) is close to zero.

\clearpage

\begin{figure}[!t]
    \centering
    \includegraphics[
        width=\linewidth,
    ]{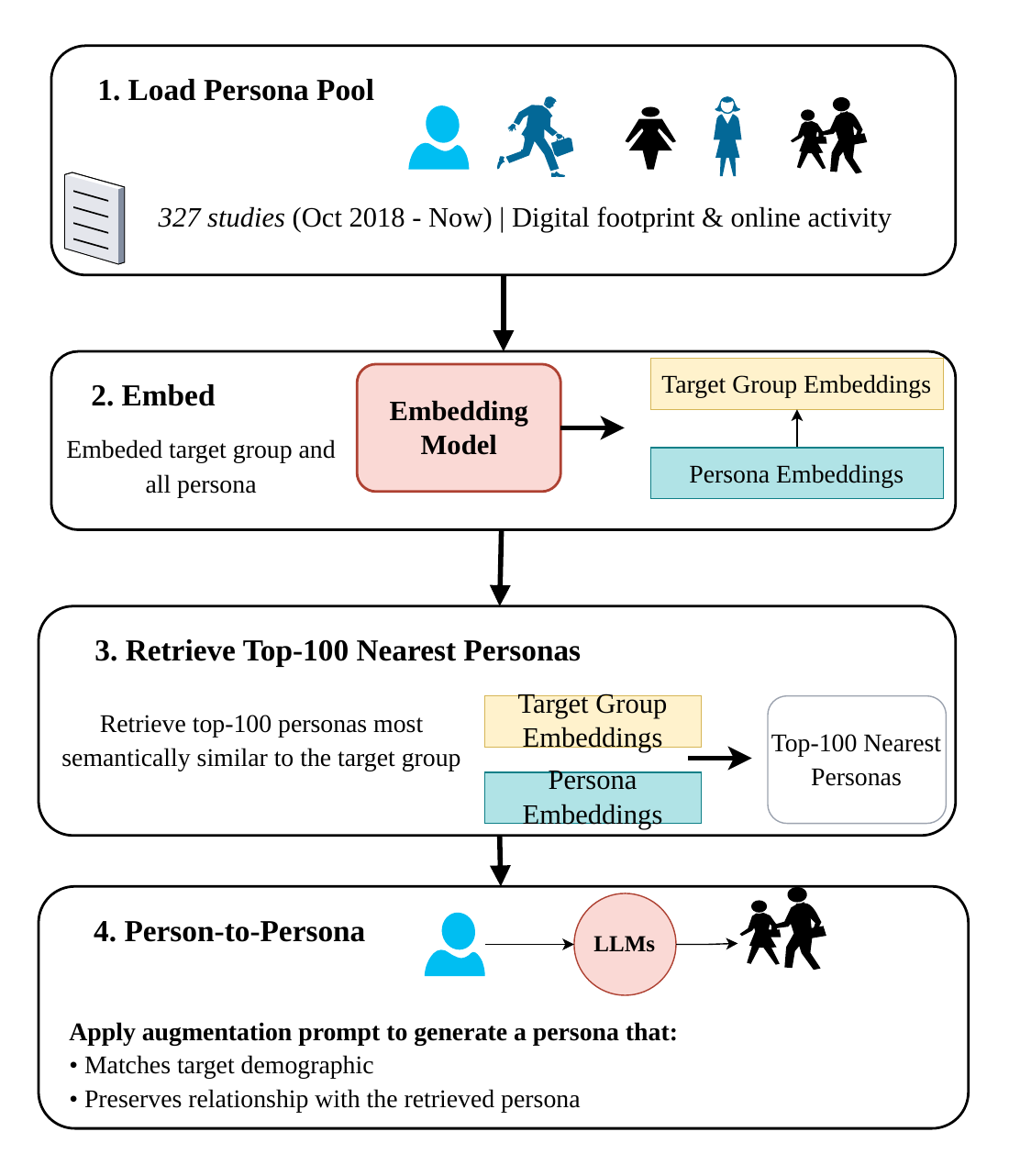}
    \caption{Persona-to-Persona Augmentation.} 
    \label{fig:persona_augmentation}
\end{figure}

\begin{figure}[t]
\centering

\begin{tcolorbox}[
    enhanced,
    width=0.95\linewidth,
    colback=gray!2,
    colframe=black,
    boxrule=0.8pt,
    arc=2mm,
    title={Persona-to-Persona Augmentation Prompt}, %
    colbacktitle=black,
    coltitle=white,
    fonttitle=\bfseries,
    left=2mm,
    right=2mm,
    top=1.5mm,
    bottom=1.5mm,
]

\textbf{Task} \\
Generate a concise persona based on the given seed persona.

\textbf{Input}

\begin{itemize}
    \item \textbf{Seed Persona:} \textit{\{seed persona\}}
    
    \item \textbf{Target Social Group}
    \begin{itemize}
        \item Dimension\textbf{:} \textit{\{dimension\} }
        \hfill (e.g., gender, social class)
        
        \item Group: \textit{\{social group name\}}
        \hfill (e.g., women, upper)
    \end{itemize}
\end{itemize}

\textbf{Requirements}

\begin{itemize}
    \item The generated persona must reflect the target dimension.
    
    \item The persona must maintain a meaningful relationship with the seed persona
    (e.g., friend, sibling, coworker).
    
    \item Keep the persona short, natural, and specific.
    
    \item Avoid generic or overly descriptive profiles.
\end{itemize}

\textbf{Output Format} \\
Return \textbf{ONLY} the generated persona.
\end{tcolorbox}
\captionof{figure}{\centering Persona-to-Persona Augmentation Prompt}
\label{fig:persona_augmentation_prompt}
\end{figure}

\begin{table}[!t]
\centering
\small
\caption{Attributes and corresponding values used for social-group construction.}
\label{tab:social_attributes}
\resizebox{\linewidth}{!}{
\begin{tabular}{ll}
\toprule
\textbf{Attribute} & \textbf{Value} \\
\midrule

\multirow{3}{*}{Social Class}
& Upper \\
& Middle \\
& Lower \\

\multirow{3}{*}{Gender}
& Male \\
& Female \\
& Non-binary \\

\multirow{4}{*}{Location}
& Urban \\
& Suburban \\
& Rural \\
& Remote / isolated \\

\multirow{5}{*}{Family and Relationships}
& Single \\
& In a relationship (not married) \\
& Married \\
& Divorced / separated \\
& Widowed \\

\multirow{5}{*}{Sexual Orientation}
& Heterosexual \\
& Homosexual \\
& Bisexual \\
& Asexual \\
& Other / prefer not to say \\

\multirow{4}{*}{Political Orientation}
& Left \\
& Center \\
& Right \\
& Apolitical / no affiliation \\

\multirow{8}{*}{Ethnicity and Race}
& Asian \\
& Black \\
& White \\
& Hispanic / Latino \\
& Middle Eastern / North African \\
& Indigenous \\
& Multiracial \\
& Other / prefer not to say \\

\multirow{6}{*}{Education}
& Primary \\
& Secondary \\
& High school graduate \\
& Vocational / technical \\
& Bachelor’s degree \\
& Postgraduate (Master/PhD) \\

\multirow{5}{*}{Income}
& Low income \\
& Lower-middle income \\
& Middle income \\
& Upper-middle income \\
& High income \\

\multirow{4}{*}{English Proficiency}
& Native speaker \\
& Fluent non-native \\
& Intermediate \\
& Basic \\

\multirow{4}{*}{Health}
& Generally healthy \\
& Chronic physical condition \\
& Disability \\
& Mental health condition \\

\multirow{7}{*}{Religion}
& Christianity \\
& Islam \\
& Hinduism \\
& Buddhism \\
& Judaism \\
& Other religion \\
& No religion (atheist/agnostic) \\

\multirow{4}{*}{Age Group}
& Teenager (13--19) \\
& Young Adult (20--35) \\
& Middle Age (36--55) \\
& Senior (56+) \\

\multirow{13}{*}{Occupation}
& Management, Business, and Financial Occupations \\
& Computer, Engineering, and Science Occupations \\
& Education, Legal, Community Service, Arts, and Media Occupations \\
& Healthcare Practitioners and Technical Occupations \\
& Service Occupations \\
& Sales and Related Occupations \\
& Office and Administrative Support Occupations \\
& Farming, Fishing, and Forestry Occupations \\
& Construction and Extraction Occupations \\
& Installation, Maintenance, and Repair Occupations \\
& Production Occupations \\
& Transportation and Material Moving Occupations \\
& Military Specific Occupations \\

\bottomrule
\end{tabular}
}
\end{table}

\clearpage
\onecolumn
\subsection{Prompts}
\label{sec:appendix_prompts}

\begin{tcolorbox}[
    enhanced,
    width=0.95\linewidth,
    colback=gray!2,
    colframe=black,
    boxrule=0.8pt,
    arc=2mm,
    title=\textbf{Take Action Prompt},
    colbacktitle=black,
    coltitle=white,
    fonttitle=\bfseries,
    left=2mm,
    right=2mm,
    top=1.5mm,
    bottom=1.5mm,
]

Question:  
\textit{\{question\}}

Instruction:  
Please answer the question for the following user profile: \textit{\{social group\}}

\textit{\{conversation history\}}

\end{tcolorbox}

\begin{tcolorbox}[
    enhanced,
    width=0.95\linewidth,
    colback=gray!2,
    colframe=black,
    boxrule=0.8pt,
    arc=2mm,
    title=\textbf{Preference Prediction Prompt},
    colbacktitle=black,
    coltitle=white,
    fonttitle=\bfseries,
    left=2mm,
    right=2mm,
    top=1.5mm,
    bottom=1.5mm,
]

You are an assistant that determines the response style of an answer.

Question:  
\textit{\{question\}}

Answer:  
\textit{\{answer\}}

Available Rubrics:  
\textit{\{rubrics\}}

Task:  
Analyze the question and identify the appropriate level for each rubric reflected in the answer.

Output Format:  
Return the result strictly as valid JSON with keys exactly matching the rubric names, and values indicating the preferred level for that rubric.

Example Output: \textit{\{example output\}}

Important:
\begin{itemize}[leftmargin=4mm]
    \item Only output JSON. Do not include explanations or additional text.
    \item Ensure all rubric keys from the provided rubrics appear in the output.
\end{itemize}

\end{tcolorbox}

\begin{tcolorbox}[
    enhanced,
    width=0.9\linewidth,
    colback=gray!2,
    colframe=black,
    boxrule=0.8pt,
    arc=2mm,
    title=\textbf{Get Universal Truth Reward},
    colbacktitle=black,
    coltitle=white,
    fonttitle=\bfseries,
    left=2mm,
    right=2mm,
    top=1.5mm,
    bottom=1.5mm,
]

\textbf{Task:}  
You are an evaluator whose task is to determine whether the model's final answer is factually correct with respect to the provided ground truth.

\textbf{Evaluation Rules:}
\begin{itemize}
    \item Evaluate \textbf{only} the final conclusion.
    \item Ignore incorrect intermediate reasoning if the final answer is correct.
    \item Semantic equivalence counts as correct.
    \item Minor wording or formatting differences should not affect correctness.
    \item If the correct answer is clearly stated, return \texttt{1}.
\end{itemize}

\textbf{Context:}

\# Question:
\textit{\{question\}}

\# Ground Truth Final Answer
\textit{\{ground\_truth\}}

\# Answer to Evaluate
\textit{\{answer\}}

\textbf{Required Output Format:}

\begin{quote}
\small
\texttt{<reasoning> optional reasoning </reasoning>}\\
\texttt{<truth\_acc>1</truth\_acc>}
\end{quote}

or

\begin{quote}
\small
\texttt{<reasoning> optional reasoning </reasoning>}\\
\texttt{<truth\_acc>0</truth\_acc>}
\end{quote}
\end{tcolorbox}

\begin{tcolorbox}[
    enhanced,
    width=0.95\linewidth,
    colback=gray!2,
    colframe=black,
    boxrule=0.8pt,
    arc=2mm,
    title=\textbf{Environment Feedback Prompt},
    colbacktitle=black,
    coltitle=white,
    fonttitle=\bfseries,
    left=2mm,
    right=2mm,
    top=1.5mm,
    bottom=1.5mm,
]

Question:  
\textit{\{question\}}

Instruction:  
You are given \textit{\{number\_agents\}} answers generated by different agents.  
Compare all answers and identify differences regarding universal truth or factual correctness in both the final conclusions and reasoning steps.

If there are inconsistencies:
\begin{itemize}
    \item Identify which statements conflict across agents.
    \item For each agent, generate concise feedback describing how its answer differs from the others.
    \item Use the following style:
    
    \textit{``While your answer claims that [statement A], other answer(s) suggest [negation or conflicting statement].''}
\end{itemize}

If all answers are universally consistent:
\begin{itemize}
    \item State that all agents agree on the same universal truth.
    \item Then provide concise suggestions for improving answer style, clarity, or helpfulness based on other agents’ responses.
\end{itemize}

Consistency Rule:
\begin{itemize}
    \item Ensure all feedbacks are mutually consistent.
    \item Do not produce contradictory judgments across agents.
\end{itemize}

Answers:  

\textit{\{answers\}}

Output Format:  

\texttt{Feedback to Agent <agent\_name>: [feedback]}  

\texttt{<split>}

\end{tcolorbox}

\begin{tcolorbox}[
    enhanced,
    width=0.95\linewidth,
    colback=gray!2,
    colframe=black,
    boxrule=0.8pt,
    arc=2mm,
    title=\textbf{Rubric Generation Prompt},
    colbacktitle=black,
    coltitle=white,
    fonttitle=\bfseries,
    left=2mm,
    right=2mm,
    top=1.5mm,
    bottom=1.5mm,
]

You are an assistant that determines the user's preferred response style based on their profile and the question.

User Profile:  
\textit{\{user\_profile\}}

Question:  
\textit{\{question\}}

Available Rubrics:  
\textit{\{rubrics\}}

Task:  
Analyze the user profile and the question, and identify the level for each rubric that the user would prefer in the answer.

Output Format:  
Return the result strictly as valid JSON with keys exactly matching the rubric names, and values indicating the preferred level for that rubric.

Example Output: \textit{\{example output\}}

Important:
\begin{itemize}[leftmargin=4mm]
    \item Only output JSON. Do not include any explanations, notes, or text outside the JSON.
    \item Ensure all rubric keys from the given rubrics are present in the output.
\end{itemize}

\end{tcolorbox}

\clearpage

\begin{table*}[t]
\centering
\small
\caption{Predefined personalization preference rubrics. Each dimension contains multiple candidate styles with corresponding descriptions and decision criteria.}
\label{tab:predefined_rubrics}
\resizebox{\linewidth}{!}{
\begin{tabular}{p{3.2cm}|p{2.2cm}|p{5.4cm}|p{5.8cm}}
\toprule

\textbf{Preference Dimension}
& \textbf{Preference Type}
& \textbf{Description}
& \textbf{Decision Rule} \\

\midrule

\multirow{3}{*}{Information Complexity}
& Basic
& Uses everyday language, avoids jargon, and explains concepts explicitly.
& If a complete beginner can understand without additional background knowledge. \\

& Intermediate
& Introduces some technical terminology while remaining accessible.
& If readers with basic familiarity can follow comfortably. \\

& Expert
& Uses advanced terminology and domain-specific concepts with minimal explanation.
& If understanding requires professional or domain expertise. \\

\midrule

\multirow{3}{*}{Structure}
& Paragraph
& Continuous prose with implicit logical flow.
& If ideas are presented naturally without explicit formatting. \\

& Bullet Points
& Information is divided into discrete independent points.
& If readability is improved through segmentation rather than sequence. \\

& Step-by-Step
& Sequential structure where each step depends on previous steps.
& If order is essential for understanding or execution. \\

\midrule

\multirow{3}{*}{Evidence Style}
& Explanation + Example
& Uses illustrative or hypothetical examples for intuition.
& If examples primarily clarify concepts rather than validate claims. \\

& Explanation + Hard Evidence
& Supports claims with measurable or verifiable evidence.
& If credibility depends on statistics, studies, or empirical findings. \\

& Explanation + Anecdotal / Expert Evidence
& Relies on authority, experience, or real-world examples.
& If credibility mainly comes from expertise or historical cases. \\

\midrule

\multirow{3}{*}{Information Density}
& Simplified
& Focuses only on key ideas while minimizing detail.
& If adding more detail would mainly increase complexity. \\

& Balanced
& Balances clarity and completeness with moderate detail.
& If the response maintains both readability and sufficient depth. \\

& Dense
& Packs large amounts of information into compact text.
& If the response maximizes information density and completeness. \\

\midrule

\multirow{3}{*}{Reasoning Style}
& Intuitive
& Relies on analogies and conceptual mental models.
& If understanding is driven mainly by intuition rather than formal logic. \\

& Analytical
& Uses systematic decomposition and logical reasoning.
& If reasoning follows explicit step-by-step logical structure. \\

& Causal Reasoning
& Explains mechanisms and cause-effect relationships.
& If explanations focus on why and how outcomes occur. \\

\midrule

\multirow{3}{*}{Tone}
& Friendly
& Conversational, approachable, and engaging.
& If the response resembles informal human conversation. \\

& Neutral
& Objective and emotionally neutral presentation.
& If the response focuses strictly on clarity and factuality. \\

& Professional
& Formal and precise communication style.
& If the tone resembles academic or workplace communication. \\

\bottomrule
\end{tabular}
}
\end{table*}

\clearpage
\subsection{Data Statistics}
\label{sec:data_statistics}
\begin{table}[t]
\centering
\small
\begin{tabular}{llr}
\toprule
\textbf{Split} & \textbf{Dataset} & \textbf{\# Samples} \\
\midrule
\multirow{4}{*}{Test}
& AIME25 & 180 \\
& StereoSet & 2352 \\
& TruthfulQA & 4902 \\
& MMLU-Pro & 420 \\
\midrule
\multirow{4}{*}{Train}
& COLM-LIMO & 817 \\
& LogicQA & 651 \\
& StereoSet & 500 \\
& SimpleQA & 300 \\
\bottomrule
\end{tabular}
\caption{Dataset statistics. The total number of training trajectories after filtering is 29,982.}
\label{tab:data_stats}
\end{table}

\begin{table}[t]
\centering
\small
\begin{tabular}{lrrrr}
\toprule
\textbf{Split} & \textbf{\# Categories} & \textbf{\# Social Groups} & \textbf{\# Personas / Group} & \textbf{\# Personas} \\
\midrule
Train & 14 & 75 & 80 & 6000 \\
Test & 2 & 6 & 20 & 120 \\
\bottomrule
\end{tabular}
\caption{Persona statistics. Due to computational resource limitations, we random a subset of social groups for evaluation.}
\label{tab:persona_stats}
\end{table}
\clearpage

\clearpage
\begin{figure}[!t]
    \centering
    \includegraphics[
        width=\linewidth,
    ]{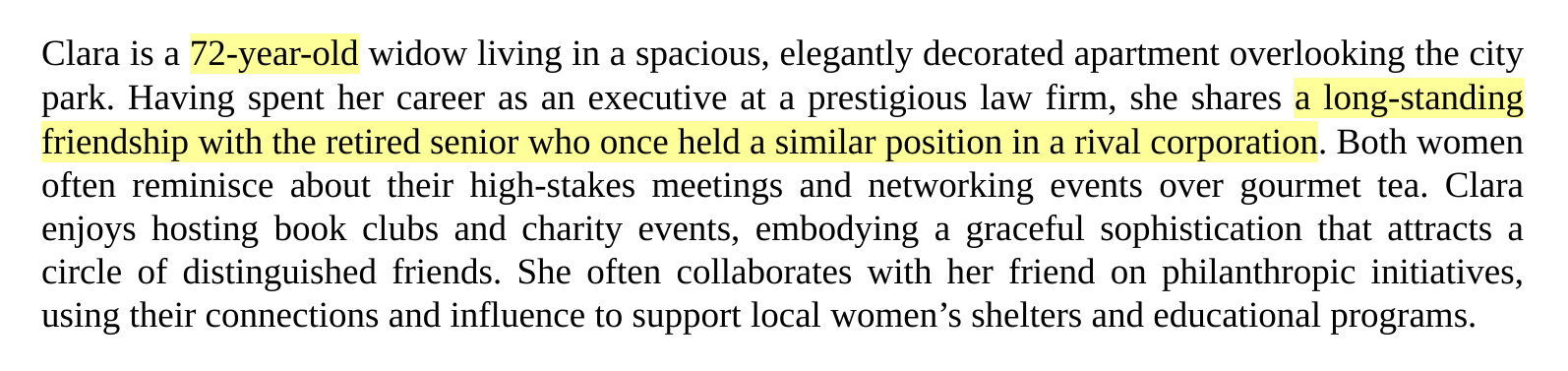}
    \caption{Persona Example.} 
    \label{fig:persona_example}
\end{figure}

\end{document}